\documentclass{article}

\usepackage{arxiv}
\usepackage{fancyhdr}
\usepackage[utf8]{inputenc} 
\usepackage[T1]{fontenc}    
\usepackage{hyperref}       
\usepackage{url}            
\usepackage{booktabs}       
\usepackage{amsfonts}       
\usepackage{nicefrac}       
\usepackage{microtype}      
\usepackage{lipsum}		
\usepackage{graphicx}
\usepackage[numbers]{natbib}
\usepackage{doi}
\usepackage{amsmath}
\usepackage{multirow}

\title{Shape-morphing programming of soft materials on complex geometries via neural operator}

\author{
	Lu Chen\textsuperscript{1,5}, 
	Gengxiang Chen\textsuperscript{1,2,5}, 
	Xu Liu\textsuperscript{3,5}, 
	Jingyan Su\textsuperscript{1}, 
	Xuhao Lyu\textsuperscript{1}, 
	Lihui Wang\textsuperscript{4}, 
	Yingguang Li\textsuperscript{1,*} \\[0.5em]
	\textsuperscript{1}College of Mechanical and Electrical Engineering, Nanjing University of Aeronautics and Astronautics, \\
	Nanjing, China \\
	\textsuperscript{2}Université Sorbonne Paris Nord, Villetaneuse, France \\
	\textsuperscript{3}School of Mechanical and Power Engineering, Nanjing Tech University, Nanjing, China \\
	\textsuperscript{4}Department of Production Engineering, KTH Royal Institute of Technology, Stockholm, Sweden \\[0.3em]
	\textsuperscript{5}These authors contributed equally: Lu Chen, Gengxiang Chen, and Xu Liu \\[0.1em]
	\textsuperscript{*}Corresponding author, Email: liyingguang@nuaa.edu.cn
}

\renewcommand{\shorttitle}{\textit{arXiv} Template}

\begin{document}
\maketitle

\begin{abstract}
Shape-morphing soft materials can enable diverse target morphologies through voxel-level material distribution design, offering significant potential for various applications. Despite progress in basic shape-morphing design with simple geometries, achieving advanced applications such as conformal implant deployment or aerodynamic morphing requires accurate and diverse morphing designs on complex geometries, which remains challenging. Here, we present a Spectral and Spatial Neural Operator (S2NO), which enables high-fidelity morphing prediction on complex geometries. S2NO effectively captures global and local morphing behaviours on irregular computational domains by integrating Laplacian eigenfunction encoding and spatial convolutions. Combining S2NO with evolutionary algorithms enables voxel-level optimisation of material distributions for shape morphing programming on various complex geometries, including irregular-boundary shapes, porous structures, and thin-walled structures. Furthermore, the neural operator’s discretisation-invariant property enables super-resolution material distribution design, further expanding the diversity and complexity of morphing design. These advancements significantly improve the efficiency and capability of programming complex shape morphing.
\end{abstract}


\section{Introduction}

\begin{figure}
	\centering
	\includegraphics[width=1.\linewidth]{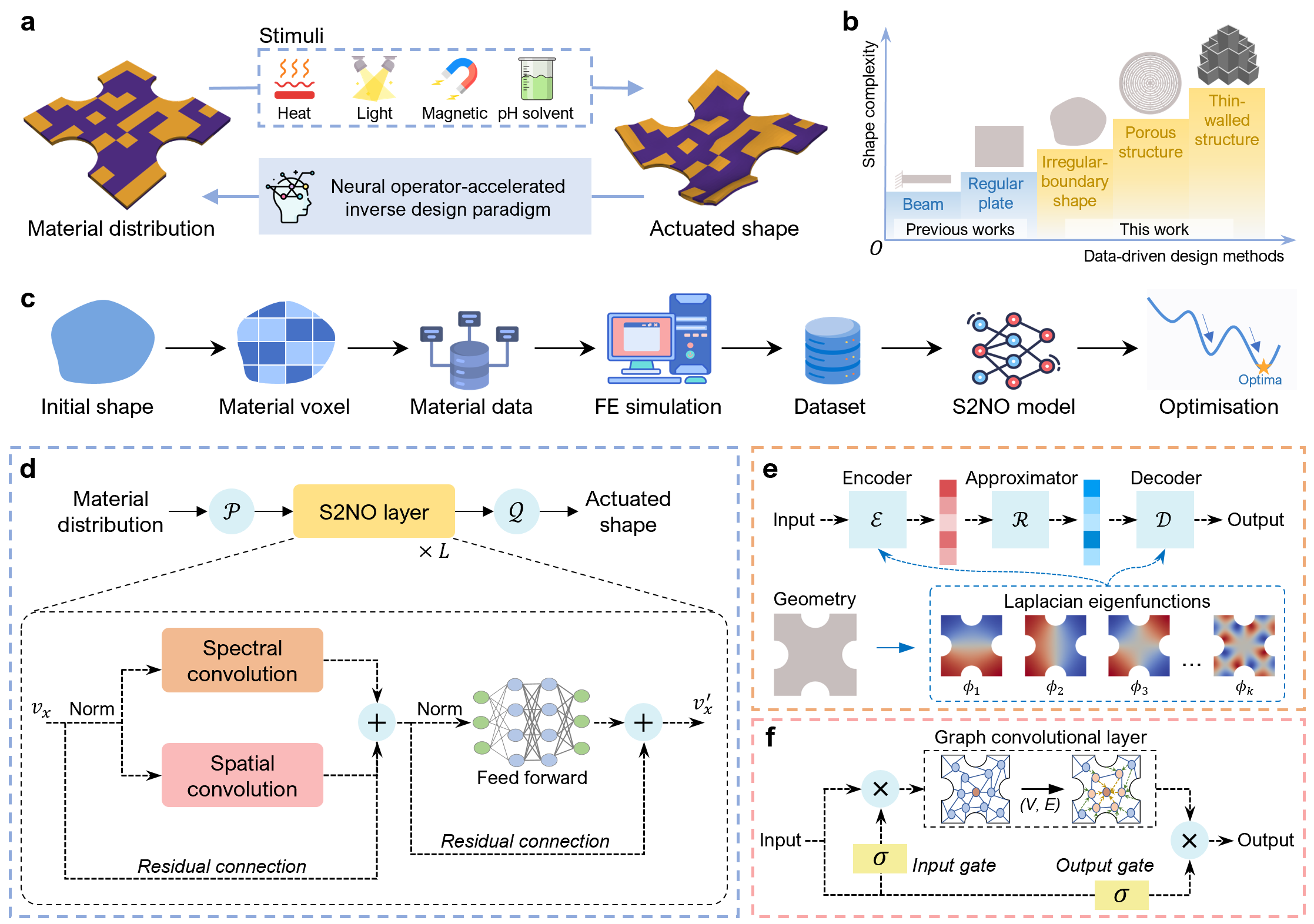}
	\caption{\textbf{Illustration of programming shape morphing via S2NO.} \textbf{a}, Shape-morphing soft materials can alter their shapes in response to external stimuli, and the desired morphing behaviour can be achieved through inverse design of material distribution. \textbf{b}, Devisable shape complexity compared to previous data-driven design studies. \textbf{c}, A complete process for the proposed design framework via S2NO. \textbf{d}, A S2NO model architecture. S2NO consists of a lifting layer $P$ to lift the channel dimension, multiple S2NO layers to learn the operator, and a projecting layer $Q$ to project the data back to the output channel dimension. Each S2NO layer comprises two key components: spectral convolution and spatial convolution. \textbf{e}, Spectral convolution is an encoder-approximator-decoder block. The encoder projects high-dimensional data into a low-dimensional frequency domain based on Laplacian eigenfunctions. Approximator learns the mappings in the frequency domain. Decoder reconstructs the learned frequency-domain data into the spatial domain. \textbf{f}, Spatial convolution consisting of a graph convolutional layer aggregating information from neighbouring nodes, an input gate, and an output gate. Two gates are designed to learn the weights of the contributions of spatial convolution.}
	\label{fig:fig1}
\end{figure}

Shape-morphing soft materials can change their shape in response to external stimuli, such as heat, light, or electric/magnetic fields (\textbf{Fig.} \ref{fig:fig1}a) \cite{wang2024performance,he2024programmable}, offering broad potential for applications in soft robotics \cite{hu2018small,bao2025real,xu2025transforming}, biomedical devices \cite{li2022soft,mao2024magnetic}, and aerospace systems \cite{liu2014shape,mahmood2023revolutionizing}. These specific morphing behaviours are programmed into the voxel-level spatial distribution of material properties, such as magnetisation \cite{meng2024programmable}, molecular orientation \cite{yang2025active}, or swelling ratio \cite{nojoomi2018bioinspired}, typically achieved through three-dimensional (3D) or four-dimensional (4D) printing \cite{gladman2016biomimetic,kim2018printing,wang2023programmable}. Although basic shape-morphing design has been achieved in various material systems with simple or regular geometries, many functional applications inherently require shape morphing on complex geometries. For instance, biomedical implants must precisely conform to the irregular surfaces of internal organs to ensure intimate contact and functional integration, while next-generation aerodynamic morphing aircraft are envisioned to actively reshape their airframe and wing surface profiles through complex deformation to meet varying flight performance requirements \cite{sun2025stimuli,yarali2024d}. Therefore, the diverse functional requirements and complex morphing environments demand shape-morphing materials capable of achieving accurate and diverse deformations on complex initial geometries. These requirements pose challenges in achieving high morphing-mode freedom and accurate modelling of morphing behaviour from high-resolution material distribution on complex geometries.

Traditional intuition-driven approaches are not ideal due to the intrinsic coupling of design variables and complex nonlinear deformation mechanisms. Finite element (FE) simulation-based methods suffer from high computational costs, thereby simplifying the high-dimensional material-deformation mapping using a few design variables, which restricts the design space and limits the diversity and accuracy of achievable morphing behaviours \cite{wang2021evolutionary,peng2023controllable,averitt2025artificial}. Although several analytical approaches, such as conformal flattening, have been proposed to derive material distributions through geometric mappings \cite{nojoomi20212d,xia2025inverse}, they are inherently restricted to conformally mappable geometries and specific material systems (e.g., swelling hydrogels) with pre-defined morphing modes \cite{sun2024machine,li2023general}. Recently, data-driven methods have emerged as a promising paradigm for programming shape morphing, offering computational efficiency and high-fidelity prediction for high-dimensional design problems \cite{raabe2023accelerating,sun2024perspective,cheng2023programming,forte2022inverse}. For example, fully connected neural networks (FCNN) for magnetic soft materials \cite{karacakol2025data}, recurrent neural networks (RNN) for active composite (AC) beams \cite{sun2022machine}, and residual networks (ResNet) for AC square plates \cite{sun2024machine}. However, these studies remain largely limited to simple regular geometries and are not applicable to learning the mapping between material distributions and deformation fields on irregular computational domains. Consequently, achieving accurate and diverse shape-morphing programming on complex geometries remains challenging.

Since both the material distribution and deformation field are functions defined on complex geometries, we propose to formulate this mapping as an operator learning problem. Therefore, in this work, we present a Spectral and Spatial Neural Operator (S2NO) that enables high-fidelity morphing prediction on complex geometries. As illustrated in \textbf{Fig.} \ref{fig:fig1}d-f, the proposed S2NO model, which integrates spectral convolutions and spatial convolutions, can capture both global and local morphing behaviours on irregular computational domains. By combining the trained S2NO with advanced optimisation algorithms, our approach can complete an exhaustive exploration across the design space, thus enabling accurate inverse design of diverse target morphologies for complex geometries (\textbf{Fig.} \ref{fig:fig1}b-c). Furthermore, the S2NO’s discretisation-invariance facilitates super-resolution material distribution design \cite{wang2023scientific,azizzadenesheli2024neural}, substantially broadening the diversity and complexity of the achievable morphing behaviours. Experiments on various complex geometries, including irregular-boundary shapes, porous structures, and thin-walled structures, demonstrate the exceptional predictive performance of the S2NO model and the robust design capabilities of our design framework. We further demonstrate the potential of our framework for designing multiple similar shape-morphing units, which can achieve more complex target shapes through modular assembly. This research provides a general framework for inverse design involving complex-geometry morphing, thereby facilitating broader applications of shape-morphing soft materials in practice.

\section{Results}
\label{sec:results}

\subsection{Verification cases and data acquisition}

To test the performance of our S2NO-driven design framework, we design a series of shape-morphing soft materials featuring various complex geometries, including irregular-boundary shapes (dart, human, stingray, and blade shapes), porous structures (dome and butterfly structures), and thin-walled structures. In this work, we demonstrate the performance of our design framework by using thermal-responsive composites as an example. This framework can easily be adapted for the design of other stimuli-responsive materials. Thermal-responsive composites, which combine active and passive materials, are a common approach for achieving shape changes \cite{xia2022responsive}. The active material responds to a heat stimulus (e.g., polylactic acid), while the passive material is insensitive to heat (e.g., thermoplastic polyurethane), resulting in a strain mismatch that induces morphing \cite{peng2023controllable,liu20234d}.

The initial shapes and material voxels for the studied cases are assigned to figures as follows: dart shape in \textbf{Fig.} \ref{fig:fig2}a, human shape in \textbf{Fig.} \ref{fig:fig2}c, stingray shape in \textbf{Fig.} \ref{fig:fig2}e, dome structure in \textbf{Fig.} \ref{fig:fig4}a, butterfly structure in \textbf{Fig.} \ref{fig:fig4}c, thin-walled structure in \textbf{Fig.} \ref{fig:fig4}e, and blade shape in \textbf{Fig.} \ref{fig:fig5}a. All shapes have a thickness of 1 mm and are subdivided into two material voxels in the thickness direction. Each material voxel is assigned an active or passive material to achieve a specific target shape. The coefficient of thermal expansion is set to 0.001 for the active material and 0 for the passive material \cite{sun2024machine,sun2022machine}. FE simulation is used to obtain the actuated shape for a given material distribution in this work (see Methods and Supplementary Figs. 1-3). To train data-driven models, we generate 55,000 random material distributions for each case in the vast design space, and the corresponding actuated shapes are obtained using FE simulations. Let $n$ be the number of discretisation points on the geometry. A material distribution and the corresponding actuated shape can be represented as $\boldsymbol{a} \in \mathbb{R}^n$ and $\boldsymbol{u} \in \mathbb{R}^{n \times 3}$, respectively. The numbers of discrete points in the seven cases are 7,385, 7,835, 6,515, 8,778, 9,525, 10,356, and 7,695, respectively. Both material distributions and actuated shapes are collected to compose the dataset, which is split into a training dataset with 50,000 data and a test dataset with 5,000 data.

\subsection{Predictive performance}

In \textbf{Table} \textbf{\ref{tab:tab1}}, we present a comparison of the results between S2NO and existing mainstream neural operator models on seven shape-morphing prediction tasks that cover various complex geometries, including four irregular-boundary shapes, two porous structures, and a thin-walled structure. Details of the model training are provided in the Methods section and Supplementary Note 1.2. The mean relative L2 error (L2), mean absolute error (MAE), and mean maximum point-to-point error (M-Max) are used to evaluate the performance of surrogate models. M-Max is the mean value of the maximum point-to-point error between predictive shapes and ground truth across all testing samples. Across all seven benchmark cases, S2NO consistently achieves state-of-the-art results for all metrics compared to previous models, including PODNN \cite{lanthaler2023operator,bhattacharya2021model}, DeepONet \cite{lu2021learning}, POD-DeepONet \cite{lu2022comprehensive}, NORM \cite{chen2024learning}, and Transolver \cite{wu2024transolver}. For the L2 metric, the promotion of S2NO over the best baseline can reach 54.75\%, 48.60\%, 33.96\%, 27.78\%, 20.98\%, 45.45\%, and 54.29\% in seven cases. For the MAE metric, S2NO gains promotions of 55.56\%, 50.00\%, 28.57\%, 30.00\%, 23.53\%, 50.00\%, and 50.00\% in seven cases. For the M-Max metric, the promotions are 52.34\%, 53.42\%, 35.29\%, 44.55\%, 22.73\%, 38.89\%, and 55.00\% in seven cases. Supplementary Figs. 4-10 provide the scatter plots of the predicted values of different models versus the actual values for all testing samples, as well as a visual comparison of random testing samples for all cases. The results predicted by our S2NO model show excellent agreement with the ground truth. These demonstrate the superiority of S2NO in predicting the shape morphing of soft materials with complex geometries.

\begin{table}[htbp]
	\centering
	\caption{Performance comparison of S2NO with the mainstream neural operator models on seven shape-morphing prediction tasks. Results are averaged over three repeated training. The (std) is the standard deviation. Best performance is highlighted in bold.}
	\label{tab:tab1}
	\footnotesize
	\begin{tabular}{@{}clllllll@{}}
		\toprule
		Case & Metric & PODNN & DeepONet & POD-DeepONet & NORM & Transolver & S2NO \\
		\midrule
		\multirow{3}{*}{Dart} 
		& L2 (\%) & 3.89 (2.5e-2) & 5.98 (1.0e-2) & 6.81 (2.3e-2) & 3.22 (1.4e-2) & 2.63 (3.1e-1) & \textbf{1.19 (6.0e-2)} \\
		& MAE (mm) & 0.25 (1.7e-3) & 0.42 (1.5e-3) & 0.51 (1.4e-3) & 0.22 (1.6e-4) & 0.18 (2.1e-2) & \textbf{0.08 (4.1e-3)} \\
		& M-Max (mm) & 1.36 (3.8e-3) & 2.37 (2.6e-3) & 2.51 (5.6e-3) & 1.16 (6.2e-3) & 1.07 (1.3e-1) & \textbf{0.51 (3.1e-2)} \\
		\midrule
		\multirow{3}{*}{Human} 
		& L2 (\%) & 2.07 (1.2e-2) & 3.74 (4.2e-2) & 3.88 (4.4e-3) & 1.32 (9.3e-3) & 1.07 (2.2e-2) & \textbf{0.55 (1.8e-2)} \\
		& MAE (mm) & 0.29 (1.9e-3) & 0.55 (5.7e-3) & 0.59 (5.0e-4) & 0.19 (1.6e-3) & 0.16 (2.2e-3) & \textbf{0.08 (3.0e-3)} \\
		& M-Max (mm) & 2.57 (4.2e-3) & 5.08 (4.6e-2) & 5.13 (1.8e-2) & 1.75 (1.1e-2) & 1.61 (3.0e-2) & \textbf{0.75 (1.9e-2)} \\
		\midrule
		\multirow{3}{*}{Stingray} 
		& L2 (\%) & 1.15 (3.6e-3) & 2.21 (9.1e-3) & 2.47 (1.7e-2) & 1.00 (3.3e-3) & 0.53 (1.7e-2) & \textbf{0.35 (1.1e-2)} \\
		& MAE (mm) & 0.14 (3.6e-4) & 0.27 (1.0e-3) & 0.33 (1.7e-3) & 0.12 (4.9e-4) & 0.07 (2.2e-3) & \textbf{0.05 (1.4e-3)} \\
		& M-Max (mm) & 1.04 (4.1e-3) & 2.54 (9.3e-3) & 2.46 (1.9e-2) & 1.29 (1.5e-2) & 0.51 (1.5e-2) & \textbf{0.33 (1.1e-2)} \\
		\midrule
		\multirow{3}{*}{Dome} 
		& L2 (\%) & 1.71 (1.5e-2) & 2.76 (5.5e-2) & 2.78 (3.0e-2) & 1.26 (1.1e-1) & 1.73 (6.3e-2) & \textbf{0.91 (2.9e-2)} \\
		& MAE (mm) & 0.11 (7.5e-4) & 0.21 (4.4e-3) & 0.22 (2.6e-3) & 0.10 (1.1e-2) & 0.14 (5.2e-3) & \textbf{0.07 (2.2e-3)} \\
		& M-Max (mm) & 1.14 (1.1e-2) & 1.87 (2.7e-2) & 1.63 (1.6e-2) & 1.14 (5.1e-1) & 1.10 (4.4e-2) & \textbf{0.61 (1.3e-2)} \\
		\midrule
		\multirow{3}{*}{Butterfly} 
		& L2 (\%) & 2.17 (1.0e-2) & 3.66 (1.8e-3) & 3.90 (1.0e-2) & 1.99 (2.3e-2) & 1.43 (2.3e-2) & \textbf{1.13 (2.0e-2)} \\
		& MAE (mm) & 0.25 (1.3e-3) & 0.45 (1.7e-3) & 0.50 (1.2e-3) & 0.23 (3.8e-3) & 0.17 (2.2e-3) & \textbf{0.13 (2.0e-3)} \\
		& M-Max (mm) & 2.00 (1.0e-2) & 3.94 (1.2e-2) & 3.97 (1.4e-2) & 2.05 (4.7e-2) & 1.54 (2.9e-2) & \textbf{1.19 (1.6e-2)} \\
		\midrule
		\multirow{3}{*}{\parbox{1.4cm}{\centering Thin-walled structure}} 
		& L2 (\%) & 0.93 (6.0e-4) & 1.40 (9.9e-3) & 1.16 (1.8e-2) & 0.22 (2.8e-3) & 0.30 (3.3e-3) & \textbf{0.12 (4.0e-3)} \\
		& MAE (mm) & 0.09 (7.5e-5) & 0.14 (1.1e-3) & 0.11 (1.8e-3) & 0.02 (3.0e-4) & 0.03 (3.4e-4) & \textbf{0.01 (4.1e-4)} \\
		& M-Max (mm) & 0.69 (7.5e-4) & 1.08 (1.1e-2) & 0.92 (1.5e-2) & 0.18 (2.1e-3) & 0.24 (2.8e-3) & \textbf{0.11 (2.2e-3)} \\
		\midrule
		\multirow{3}{*}{Blade} 
		& L2 (\%) & 1.00 (3.1e-3) & 1.88 (1.2e-3) & 1.82 (4.5e-3) & 0.53 (4.5e-2) & 0.35 (1.5e-2) & \textbf{0.16 (4.6e-3)} \\
		& MAE (mm) & 0.11 (3.7e-4) & 0.23 (1.4e-2) & 0.23 (7.2e-4) & 0.05 (4.5e-3) & 0.04 (1.4e-3) & \textbf{0.02 (2.3e-4)} \\
		& M-Max (mm) & 0.87 (5.8e-3) & 1.86 (7.7e-2) & 1.76 (3.3e-3) & 0.56 (5.0e-2) & 0.40 (1.9e-2) & \textbf{0.18 (3.7e-3)} \\
		\bottomrule
	\end{tabular}
\end{table}

\subsection{Programming shape morphing for irregular-boundary shapes}

\begin{figure}
	\centering
	\includegraphics[width=1.0\linewidth]{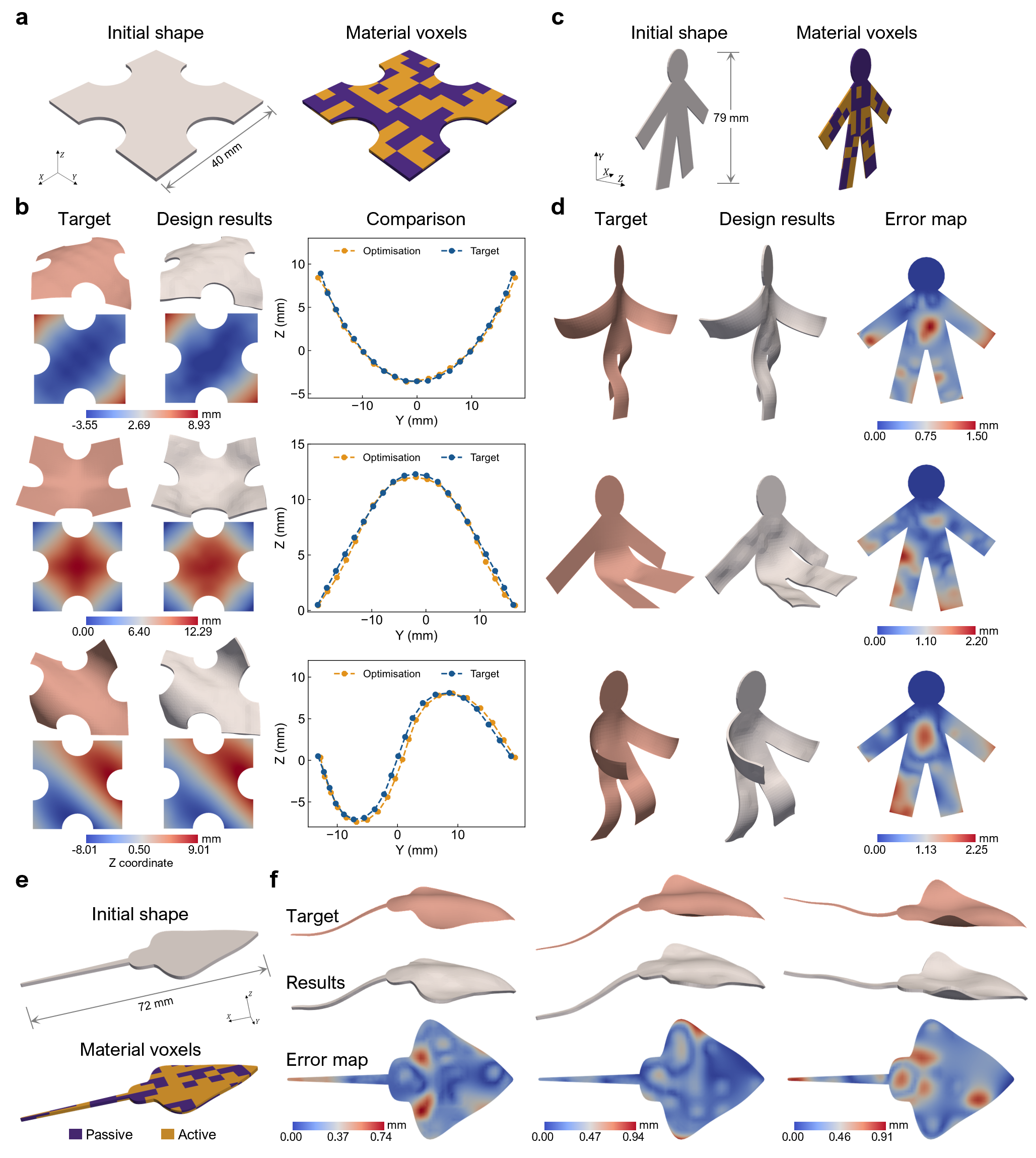}
	\caption{\textbf{Programming shape morphing for soft materials with irregular-boundary shapes.} \textbf{a}, \textbf{c}, \textbf{e}, Initial shapes and material voxel schematics for the dart case, human case, and stingray case, respectively. \textbf{b}, Target surfaces, design results, and quantitative comparison for achieving bow-shaped, parabolic, and sine function-like surfaces. \textbf{d}, Target surfaces, design results, and error maps for mimicking human behaviour, including walking, sitting, and horse stance. \textbf{f}, Target surfaces, design results, and error maps for reconstructing the swimming behaviour of the stingray with soft materials. The values in the error maps represent the point-to-point distance between the target shape and the designed shape.}
	\label{fig:fig2}
\end{figure}

This section validates the effectiveness of the S2NO-driven design framework by using it to programme the shape morphing of three complex geometries with irregular boundaries: a dart-like case, a human-like case, and a stingray-like case. \textbf{Fig.} \ref{fig:fig2}a shows the initial shape and the material voxels of the dart case. The shape has been divided into 152 material voxels to enable various shape changes, as illustrated in \textbf{Fig.} \ref{fig:fig2}b. The first target surface is bow-shaped, the second is parabolic, and the third is a surface resembling a sine function. These three targets become increasingly complex and cannot be easily achieved through intuitive designs. Meanwhile, the shapes designed using our approach and their corresponding Z-coordinate distributions are highly consistent with the targets (optimal material distributions and error maps in Supplementary Fig. 11). Comparing the coordinate variation trends along the diagonal between the target and designed shapes once again confirms this consistency.

\textbf{Fig.} \ref{fig:fig2}c illustrates the initial shape of the human case and its corresponding material voxels (129 in total). To demonstrate the flexibility of the S2NO-driven design framework, various shapes that mimic human behaviour are used as targets, including walking, sitting, and horse stance. \textbf{Fig.} \ref{fig:fig2}d shows the target shape, the design results, and the error map. Supplementary Fig. 12 provides the optimal material distributions. The arm and leg segments of these target shapes exhibit significantly different morphological changes. As can be seen, the design perfectly reproduces the target’s complex morphology, with maximum error values of just 1.90\%, 2.78\%, and 2.85\% relative to the height (79 mm) for the three target shapes. 

In the last example, we consider a stingray-inspired case, whose extensive body and flexible tail are capable of producing a variety of complex morphologies. \textbf{Fig.} \ref{fig:fig2}e shows the initial shape of the stingray case and its corresponding 160 material voxels. We design a series of shapes to replicate the sequential movements involved in a stingray’s swimming process (see \textbf{Fig.} \ref{fig:fig2}f and Supplementary Fig. 13). The edges on both sides of the body deform upwards, the central region forms a convex surface, and the tail curves and swings. \textbf{Fig.} \ref{fig:fig2}f demonstrates that our design approach produces shapes that closely match the target shapes for both the body and tail sections. The error maps reveal maximum errors of just 1.03\%, 1.31\%, and 1.26\% relative to the length (72 mm) across the three target morphologies.

These results substantiate that the S2NO-driven design framework can enable the accurate design of diverse and complex shape morphing for irregular-boundary initial shapes. The capability to accurately replicate complex motion morphologies also demonstrates the potential of our approach for application in bionic soft robotics.

\subsection{Super-resolution material distribution design}

High-resolution material voxel allows for greater design freedom and enables more complex shape-morphing behaviours. Traditional data-driven design methods are based on fixed material voxel sizes and data discretisation, and refining the material distribution resolution requires retraining the prediction model. This involves generating vast amounts of training data, given the enormous number of combinations in the design space for high-resolution material voxel. In contrast, our S2NO is a discretisation-invariant neural operator model that can predict shape morphing directly for unseen high-resolution material voxels. Furthermore, fine-tuning the model using minimal data can significantly improve the prediction accuracy, thereby supporting subsequent inverse design processes.

\begin{figure}
	\centering
	\includegraphics[width=1.\linewidth]{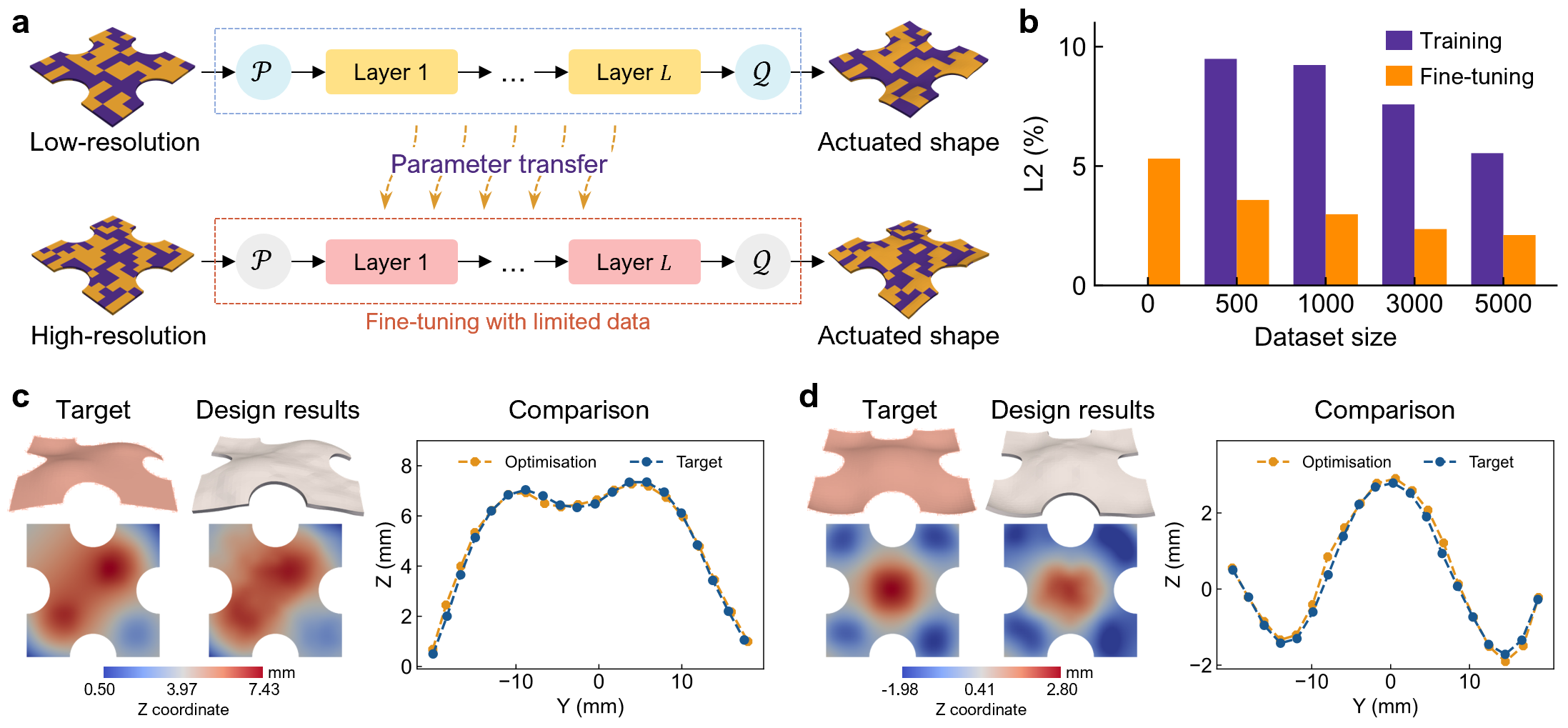}
	\caption{\textbf{Super-resolution material distribution design.} \textbf{a}, Model fine-tuning framework for transitioning from low to high material distribution resolution. \textbf{b}, Performance comparison of model fine-tuning versus direct training from scratch on limited data and direct predictions from the low-resolution model (Dataset size is 0). \textbf{c}, \textbf{d}, Target surfaces, design results, and their quantitative comparisons for achieving two complex surfaces.}
	\label{fig:fig3}
\end{figure}

\textbf{Fig.} \ref{fig:fig3}a illustrates the model fine-tuning framework across material distribution resolutions, as well as two dart shapes at different resolutions. The low-resolution case, which is identical to that in Section 2.3, uses 152 material voxels. The high-resolution case uses 290 material voxels, representing a design space $2^{138} (\approx 3 \times 10^{41})$ times larger than the low-resolution configuration. We generated 10,000 data for high-resolution material voxel via FE simulation, with 5,000 used for model fine-tuning and the remaining 5,000 for testing. To demonstrate the effectiveness of fine-tuning, \textbf{Fig.} \ref{fig:fig3}b, Supplementary Fig. 14, and Supplementary Table 2 compare the fine-tuned model against two baselines: direct prediction using the low-resolution model (Dataset size is 0) and training from scratch on limited high-resolution material voxel data. The direct prediction yields an L2 of 5.31\%, a MAE of 0.35 mm, and a M-Max of 1.86 mm—performance that surpasses even that of a model trained on 5,000 data. This indicates a zero-shot generalisation capability of the low-resolution model on unseen high-resolution voxel data. The substantial improvement achieved through fine-tuning, relative to training from scratch, demonstrates that this inherent generalisation can be significantly enhanced with only a small dataset of high-resolution voxels. \textbf{Fig.} \ref{fig:fig3}c-d and Supplementary Fig. 15 show the design results based on the prediction model fine-tuned with 5,000 data. Two more complex target surfaces than those in Section 2.3 are selected to validate the shape-morphing capability of the high-resolution material voxel. The vast design space of high-dimensional design variables also poses greater challenges to optimisation algorithms. To tackle this challenge, we propose a multi-resolution optimisation strategy that progresses from coarse to fine scales (Supplementary Fig. 16). The design results align well with the target surfaces in both overall trends and local details, demonstrating the effectiveness of S2NO model fine-tuning and the capability of our design approach.

\begin{figure}
	\centering
	\includegraphics[width=1.\linewidth]{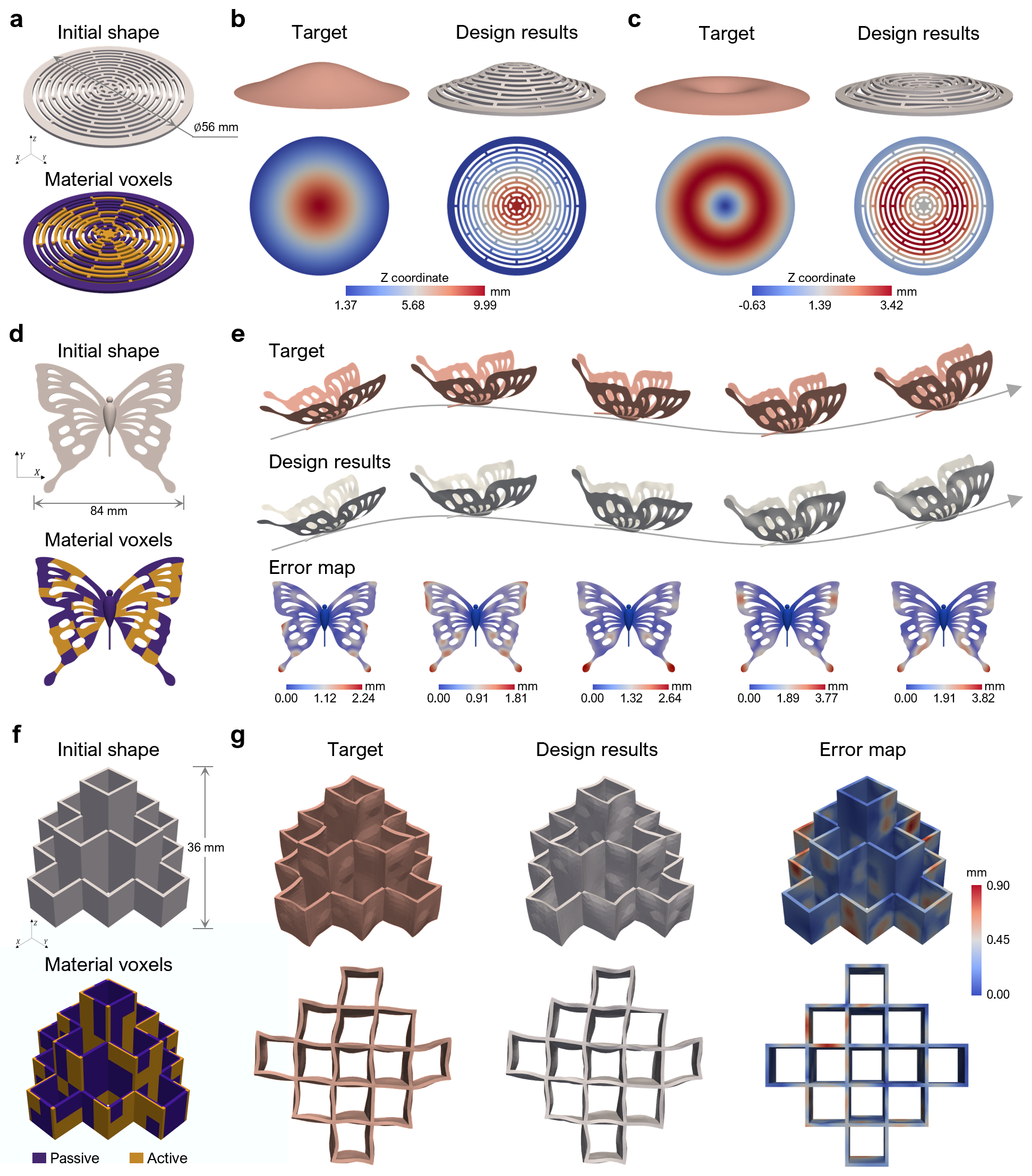}
	\caption{\textbf{Programming shape morphing for soft materials with complex-topology structures.}  \textbf{a}, \textbf{d}, \textbf{f}, Initial shapes and material voxel schematics for the dome case, butterfly case, and thin-walled structure case, respectively. \textbf{b}, \textbf{c}, Target surfaces, design results, and their Z-coordinate distributions to reconstruct a hat-like surface (\textbf{b}) and a volcano-like surface (\textbf{c}). \textbf{e}, Target surfaces, design results, and error maps for approximating five different butterfly morphologies during flight. \textbf{g}, Target surfaces, design results, and error maps for achieving a complex target shape of 3D thin-walled structure. The values in the error maps represent the point-to-point distance between the target shape and the designed shape.}
	\label{fig:fig4}
\end{figure}

\subsection{Programming shape morphing for complex-topology structures}

To further demonstrate the capability of our approach in achieving shape-morphing designs for complex geometries, we validate the performance of our S2NO-driven design framework on complex-topology structures, including two porous structures (a dome-like structure and a butterflylike structure) and a thin-walled structure. \textbf{Fig.} \ref{fig:fig4}a shows the initial shape of the dome case and its material voxels. This structure comprises 264 designable material voxels and features a circular edge with fixed constraints, allowing for the reconstruction of various concave and convex surfaces. \textbf{Fig.} \ref{fig:fig4}b-c shows the inverse design results for a hat-like surface and a volcano-like surface defined by the function $z_i^{\prime}=z_i+f\left(r_i\right) \times g\left(r_i\right)$, where $r_i=\left\|\left(x_i, y_i\right)\right\|_2, f\left(r_i\right)=A_p \exp \left\{-\left(r_i-28 \beta\right)^2 /\right. \left.\left(2 \sigma_p^2\right)\right\}+A_d \exp \left\{-r_i^2 /\left(2 \sigma_d^2\right)\right\}$, and $g\left(r_i\right)=1-\alpha\left(r_i / 28\right)^2$. The parameters $A_p, \beta, \sigma_p, A_d, \sigma_d$, and $\alpha$ for the two targets are $[10,0,12.04,0,0,0]$ and $[5,0.5,8.4,-2,8.4,1]$, respectively. Supplementary Fig. 17 provides the optimal material distributions and the error maps between the target and design. The comparison of the design results with the target surfaces demonstrates that our design strategy can provide an accurate reconstruction for complex concave and convex surfaces.

Drawing inspiration from the flight behaviour of butterflies, we designed a butterfly-like shape (\textbf{Fig.} \ref{fig:fig4}d) that mimics a variety of flight morphologies. Since butterflies are symmetrical, we use one side of their wings as the basis for the design, subdividing it into 98 material voxels. \textbf{Fig.} \ref{fig:fig4}e and Supplementary Fig. 18 provide the results of our inverse design approach for achieving five different butterfly morphologies. The design accurately captures the amplitude of the wingbeats across various flight postures, producing a remarkably realistic representation of the target shapes. The predominantly blue and cyan error map visually demonstrates the design’s high accuracy.

Finally, we demonstrate the universality of our approach by applying it to the design of a 3D thin-walled structure, which is scarcely explored by existing shape-morphing design works. \textbf{Fig.} \ref{fig:fig4}f shows the initial shape and the material voxels for the thin-walled structure. The interior of the structure is composed of passive materials, while the exterior is divided into 224 designable material voxels. We consider a target shape generated by a randomly designated material distribution. The inverse design results are shown in \textbf{Fig.} \ref{fig:fig4}g and Supplementary Fig. 19. With a maximum error value of only 0.90 mm relative to the target, the design results provide an excellent agreement with the target shape.

The above results demonstrate the capability of our S2NO-driven design framework to achieve a variety of desired shape-morphing behaviours of complex-topology structures. This is a highly challenging task for existing data-driven strategies due to the complexity of shape-morphing prediction.

\subsection{Design for modularly assembled shape-morphing structures}

\begin{figure}
	\centering
	\includegraphics[width=1.0\linewidth]{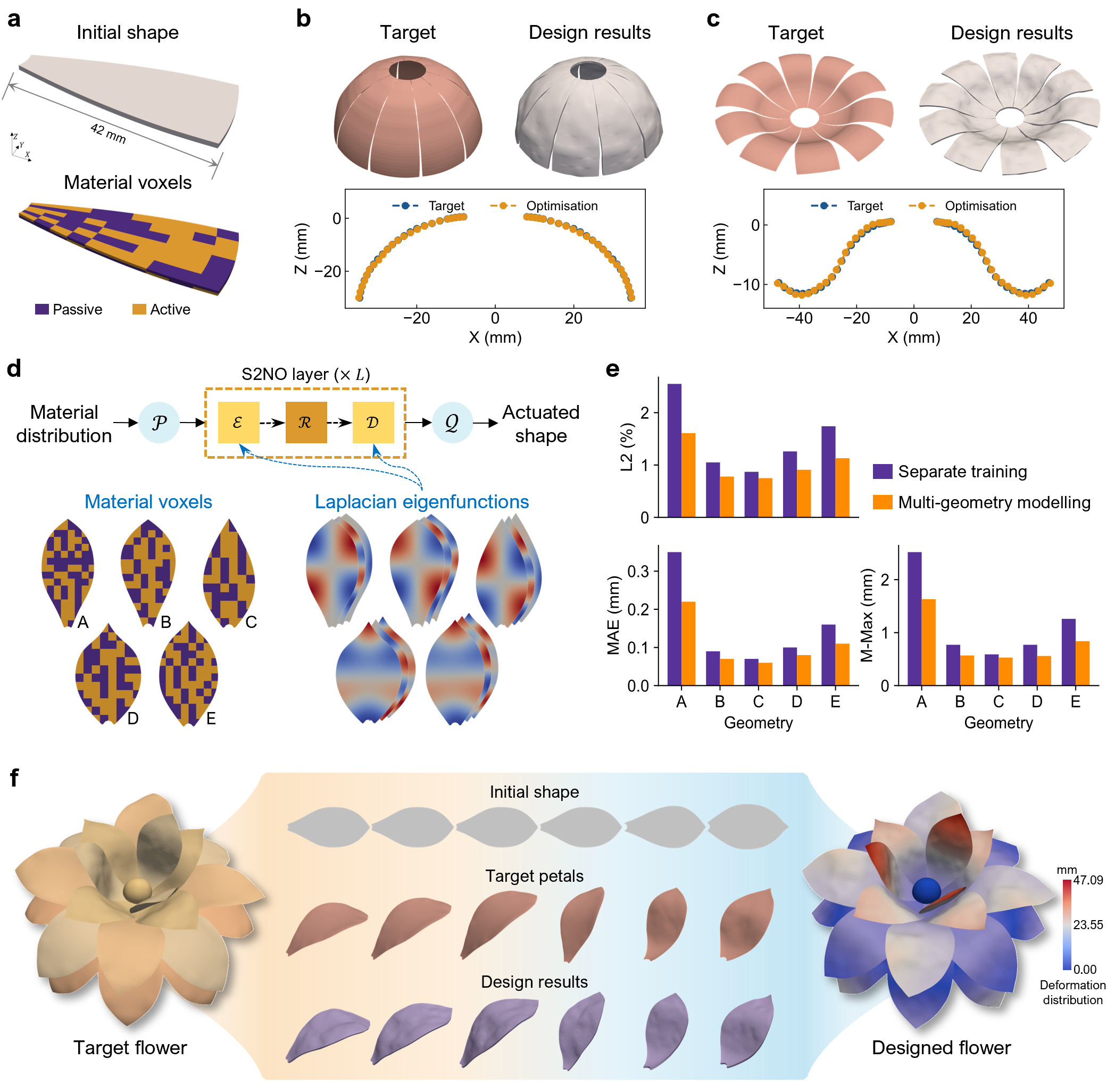}
	\caption{\textbf{Design for the modular assembled shape-morphing structures.} \textbf{a}, Initial shape of the blade case and its material voxel schematics. \textbf{b}, \textbf{c}, Target surfaces, design results, and quantitative comparison for achieving a hemispherical surface (\textbf{b}) and a morning glory-like surface (\textbf{c}) using a circular array of ten blades. \textbf{d}, S2NO-based multi-geometry modelling architecture, material voxels of five petals, and their Laplacian eigenfunctions. Model parameters of S2NO can be shared across different geometries and their Laplacian eigenfunctions. \textbf{e}, Comparison of multi-geometry modelling results for five petals with those from separate training. \textbf{f}, Design results of a hippeastrum-like flower. Initial shapes, target petals, and design results for achieving six target petal surfaces with three distinct initial geometries (A, B, and E) are provided. The colour map of the designed flower represents the deformation distribution of each petal relative to the initial shape.}
	\label{fig:fig5}
\end{figure}

The modular assembly of elementary shape-morphing units is a promising approach for constructing complex target shapes, offering key advantages in terms of manufacturability and transportability. This is particularly appealing for in-space manufacturing, as flat configurations can be launched in a highly compact form and then deployed into their functional three-dimensional shapes as required. These shape-morphing units can comprise multiple identical or similar shape-morphing structures. Here, we demonstrate the potential of the S2NO-driven design framework in such scenarios. \textbf{Fig.} \ref{fig:fig5}a shows the initial shape and the corresponding material voxels (120 in total) for the blade case. Applying a fixed constraint to the left end of the shape primarily causes it to undergo bending deformation. \textbf{Fig.} \ref{fig:fig5}b illustrates a hemispherical target surface and the corresponding design results using a circular array of ten blades. Supplementary Fig. 20 shows the optimal material distributions and the error maps. The maximum deformation value from the initial flat shape to the hemispherical target shape is 34 mm, whereas the maximum side length of the initial blade component is only 42 mm. This good consistency with the target surface demonstrates the capability of our S2NO-driven design framework to produce extreme shape-morphing behaviour. Additionally, we conduct an inverse design for another morning glory-like surface, as shown in \textbf{Fig.} \ref{fig:fig5}c. The deformation trend of a single blade resembles a sigmoid function. Once again, the design results are quantitatively consistent with the target surface.

Another type of modular assembly comprises multiple similar yet geometrically distinct shape-morphing units, such as the different petals of a hippeastrum-like flower (\textbf{Fig.} \ref{fig:fig5}f). Conventional data-driven design approaches lack the capacity to generalise across multiple geometries. This inability compels the collection of large-scale datasets and the development of separate prediction models for each shape-morphing unit, resulting in a costly and time-consuming process. In contrast, our design framework can exploit the discretisation-invariant nature of S2NO to share parameters across different geometries and their Laplacian eigenfunctions, thereby enabling multi-geometry modelling (see Methods). Through data sharing across different geometries, multi-geometry modelling enables the model training for each shape-morphing unit using reduced data requirements. \textbf{Fig.} \ref{fig:fig5}d illustrates the network architecture for S2NO-driven multi-geometry modelling, along with the shape, material voxels, and Laplacian eigenfunctions for five petals (shape size in Supplementary Fig. 3). Different petals employ their respective Laplacian eigenfunctions within the encoder and decoder of the spectral convolution. The material voxel counts for the five petals (A-E) are 180, 160, 120, 152, and 164, respectively. We generate 15,000 data for each petal via FE simulation, with 10,000 data used for training and 5,000 for testing. \textbf{Fig.} \ref{fig:fig5}e and Supplementary Table 3 compare the results of training each geometry separately with those of multi-geometry modelling. Multi-geometry modelling outperforms single-geometry training across all error metrics (L2, MAE, and M-Max) for the five petals, with particularly notable gains observed in Petal A, where all three metrics showed enhancements exceeding 35\%.

Next, we use the prediction model trained on multi-geometry data to perform inverse design on the target flower, as shown in \textbf{Fig.} \ref{fig:fig5}f and Supplementary Fig. 21. The initial shapes contain three distinct geometries (A, B, and E) to realise the six target petal surfaces. The design results for the six petals and the entire flower demonstrate a high degree of conformity with the targets. The colour map of the designed flower shows the deformation distribution of each petal relative to the initial shape. This emphasises the importance of accurately capturing shape-morphing behaviours across different magnitudes in floral design problems. These results demonstrate the capability of our design strategy based on S2NO-driven multi-geometry modelling.

\section{Discussion}

Achieving shape-morphing programming on complex geometries is necessary for realising advanced applications involving diverse functional requirements and complex morphing environments. In this work, we propose a neural-operator-driven design framework that enables accurate and diverse shape-morphing programming of soft materials on complex geometries. The mapping from the material property field to the deformation field is formulated as an operator learning problem. The proposed Spectral and Spatial Neural Operator (S2NO) captures both global and local morphing behaviours on irregular computational domains, enabling high-fidelity shape-morphing prediction for soft materials with complex geometries. The efficient forward prediction of S2NO empowers evolutionary algorithms to thoroughly explore the design space, thereby achieving accurate and diverse morphing designs.

Experiment results on various complex geometries, including irregular-boundary shapes, porous structures, and thin-walled structures, demonstrate the exceptional predictive performance of S2NO and the robust design capability of the S2NO-driven design framework. The precise design outcomes for a wide range of 3D shape morphing showcase the promising potential of our framework in fields such as soft robotics, biomimetic devices, and shape-morphing metamaterials. We also highlight the discretisation-invariant nature of neural operators as a key advantage for super-resolution material distribution design, which can expand the complexity and diversity of the morphing design space at minimal data cost. Furthermore, the S2NO-based multi-geometry modelling enables the training of a unified forward prediction model that generalises across similar geometries, improving data efficiency and enhancing generalisation capability beyond single-geometry training. While this study demonstrates a proof-of-concept using temperature-responsive composites, the design approach is generalisable to a broad range of programmable materials and external stimuli. Future research will further explore the potential of neural-operator-driven design in addressing real-world morphing challenges involving complex-geometry programming in biomedical devices and next-generation morphing aircraft.

\section{Methods}

\subsection{FE simulation}
\label{sect:4.1}

We use ABAQUS 2020 to conduct FE simulations to obtain the actuated shapes of shape-morphing soft materials with thermal-responsive composites. The active and passive materials involved in the thermal-responsive composites are modelled using the incompressible neo-Hookean model with the same parameters (C10 = 0.1 and D1 = 0.002). The coefficient of thermal expansion is set to 0.001 for active materials and to 0 for passive materials \cite{sun2024machine,sun2022machine}. A temperature increase of 80°C is applied to the thin-walled structure to induce shape morphing, while a temperature increase of 60°C is applied to all other cases. The mesh element type is set to C3D8H. The shape, material voxels, and boundary conditions for all cases are provided in Supplementary Figs. 1-3. To generate a large volume of data, we use Python to randomly generate material distributions and write them into INP files, which are subsequently submitted automatically for Abaqus computation.

\subsection{S2NO model}
\label{sect:4.2}

S2NO model follows the classical neural operator architecture \cite{kovachki2023neural,li2020fourier}, comprising a lifting layer $P$, multiple stacked S2NO layers $\mathcal{L}_{i} (i=1,…,L)$, and a projecting layer $Q$ (\textbf{Fig.} \ref{fig:fig1}d). The lifting layer lifts the input data to a higher-dimensional channel space using a pointwise neural network. The S2NO layer learns the operator. The projecting layer transforms the data back to the initial channel space. The S2NO structure can be represented as

\begin{equation}
    \mathcal{G}_\theta=Q \circ \mathcal{L}_L \circ \mathcal{L}_{L-1} \circ \ldots \circ \mathcal{L}_1 \circ P.
\end{equation}

Denote the input data for the layer $\mathcal{L}_i$ as $\boldsymbol{v}_x \in \mathbb{R}^{n \times d_c}$, where $n$ is the spatial dimension of data, $d_c$ is the channel dimension after lifting. The output of the layer $\mathcal{L}_i$ is formulated as

\begin{equation}
	\mathcal{L}_i\left(\boldsymbol{v}_x\right)=\text{FeedForward}\left(\text{LayerNorm}\left(\boldsymbol{v}_x^{m i d}\right)\right)+\boldsymbol{v}_x^{m i d},
\end{equation}

\begin{equation}
	\boldsymbol{v}_x^{m i d}=\text {Convolution}\left(\text {LayerNorm}\left(\boldsymbol{v}_x\right)\right)+\boldsymbol{v}_x,
\end{equation}
where feedforward is a pointwise neural network, the convolution operation includes spectral convolution and spatial convolution (\textbf{Fig.} \ref{fig:fig1}e-f), which capture global and local morphing behaviours, respectively. Spectral convolution first maps high-dimensional data to a low-dimensional frequency domain based on spectral basis functions (i.e., the encoder), learning mappings between low-dimensional vectors in the frequency domain (i.e., the approximator). The learned frequency-domain data is then reconstructed into the spatial domain (i.e., the decoder). To learn operators on complex geometries, we employ the Laplacian eigenfunctions of the geometry as the spectral basis functions (solved using the LaPy Python library) \cite{reuter2006laplace}. Spatial convolution is implemented via graph convolutional layers, which capture local features by aggregating information from neighbouring nodes. We use sigmoid activation functions to design input and output gates, enabling the judicious fusion of local and global features. For further details of S2NO, see the Supplementary Note 1.1.

\subsection{Model training}

The training dataset $\mathcal{D}=\left\{\left(\boldsymbol{a}_i, \boldsymbol{u}_i\right)\right\}_{i=1}^N$ consists of pairs of material distributions $\boldsymbol{a}_i \in \mathbb{R}^n$ and the corresponding actuated shape $\boldsymbol{u}_i \in \mathbb{R}^{n \times 3}$, where $N$ is the dataset size and $n$ is the number of discretisation points on the geometry . With the dataset, we aim to learning a mapping model $\mathcal{G}_{\boldsymbol{\theta}}: \boldsymbol{a} \mapsto \boldsymbol{u}$ for fast prediction of actuated shape. The learning process is mathematically formulated as

\begin{equation}
	\boldsymbol{\theta}^* \in \underset{\boldsymbol{\theta}}{\arg \min } \frac{1}{N} \sum_{i=1}^N \frac{\left\|\boldsymbol{u}_i-\mathcal{G}_{\boldsymbol{\theta}}\left(\boldsymbol{a}_i\right)\right\|_2}{\left\|\boldsymbol{u}_i\right\|_2}.
\end{equation}

For the S2NO model, we set the number of S2NO layers is 8, the channel dimension after lifting is 128, the number of Laplacian eigenfunctions is 128, the lifting layer to be a linear transform, the projecting layer to be a single-hidden-layer neural network with 128 hidden neurons, and the nonlinear activation function to be GeLU. Regarding the hyperparameters, the initial learning rate is set to 0.001, the batch size during training is set to 16, and the number of epochs is set to 500. We use the adaptive moment estimation with weight decay (AdamW) as an optimiser and the one cycle learning rate policy (OneCycleLR) as a scheduler to train the model. The training and testing are conducted on an NVIDIA A100 GPU (40G). Details on implementation for S2NO and other models (PODNN, DeepONet, POD-DeepONet, NORM, and Transolver) are provided in Supplementary Note 1.2.

\subsection{S2NO-based model fine-tuning for high-resolution material voxel}

High-resolution material voxel offers greater design freedom, yet requires vast amounts of training data to ensure the generalisability and predictive accuracy of the data-driven model. To reduce data acquisition costs, we fine-tune the existing S2NO model for low-resolution material voxels to efficiently establish a prediction model for high-resolution material voxels with limited data. Consider an S2NO model $\mathcal{G}_{\boldsymbol{\theta}_l}: \boldsymbol{a}^l \mapsto \boldsymbol{u}^l$ trained with the dataset $\mathcal{D}_L=\left\{\left(\boldsymbol{a}_i^l, \boldsymbol{u}_i^l\right)\right\}_{i=1}^{N_l}$ for the low-resolution material voxel, where $\boldsymbol{a}_i^l \in \mathbb{R}^{n_l}$ and $\boldsymbol{u}_i^l \in \mathbb{R}^{n_l \times 3}$. There also exists a very small dataset $\mathcal{D}_H=\left\{\left(\boldsymbol{a}_i^h, \boldsymbol{u}_i^h\right)\right\}_{i=1}^{N_h} \left(N_h \ll N_l\right)$ for high-resolution material voxel, where $\boldsymbol{a}_i^h \in \mathbb{R}^{n_h}$ and $\boldsymbol{u}_i^h \in \mathbb{R}^{n_h \times 3}$. $n_l$ and $n_h$ are dimensions of data discretisation, which typically differ for different resolutions. Owing to the discretisation-invariance of the S2NO model, the trained model $\mathcal{G}_{\boldsymbol{\theta}_l}$ can be directly applied to predict the output for the input $\boldsymbol{a}_i^h$. However, since the design space for highresolution material voxel is much larger than that for low-resolution, using the model $\mathcal{G}_{\boldsymbol{\theta}_l}$ to predictthe actuated shapes for the high-resolution material voxel may produce mediocre results. Consequently, we fine-tune the parameters of the model $\mathcal{G}_{\boldsymbol{\theta}_l}$ using the dataset $\mathcal{D}_H$ to obtain a new S2NO model $\mathcal{G}_{\boldsymbol{\theta}_h}:\boldsymbol{a}^h \mapsto \boldsymbol{u}^h$ with high predictive performance for high-resolution material voxel scenarios. The learning process is mathematically formulated as

\begin{equation}
	\boldsymbol{\theta}_h^* \in \underset{\boldsymbol{\theta}_h}{\arg \min } \frac{1}{N_h} \sum_{i=1}^{N_h} \frac{\left\|\boldsymbol{u}_i^h-\mathcal{G}_{\boldsymbol{\theta}_h}\left(\boldsymbol{a}_i^h\right)\right\|_2}{\left\|\boldsymbol{u}_i^h\right\|_2}, \boldsymbol{\theta}_h^0=\boldsymbol{\theta}_l,
\end{equation}
where $\boldsymbol{\theta}_h^0=\boldsymbol{\theta}_l$ represents that the initial values of parameter $\boldsymbol{\theta}_h$ are set to $\boldsymbol{\theta}_l$. The S2NO models $\mathcal{G}_{\boldsymbol{\theta}_h}$ and $\mathcal{G}_{\boldsymbol{\theta}_l}$ have the same network architectures. Moreover, $\mathcal{G}_{\boldsymbol{\theta}_h}$ and $\mathcal{G}_{\boldsymbol{\theta}_l}$ are established on the same geometry, and the Laplacian eigenfunctions of the geometry are also shareable. During execution, since the Laplacian eigenfunctions of complex geometry can only be discretely solved, we separately downsample based on the corresponding data discretisations from a set of Laplacian eigenfunctions for the models $\mathcal{G}_{\boldsymbol{\theta}_h}$ and $\mathcal{G}_{\boldsymbol{\theta}_l}$.

\subsection{S2NO-based multi-geometry modelling}

Unlike training models separately for each shape-morphing structure, multi-geometry modelling involves training a common model on multiple similar structures (akin to multi-task learning). This improves the model's generalisation performance and reduces data requirements. Consider $M$ shape-morphing structures possessing similar geometries $\Omega_i (i=1, \ldots, M)$. Assume that for each geometries $\Omega_i$ there is a training dataset $\mathcal{D}_i=\left\{\left(\boldsymbol{a}_j^i, \boldsymbol{u}_j^i\right)\right\}_{j=1}^{N_i}$, where $\boldsymbol{a}_j^i \in \mathbb{R}^{n_i}$ and $\boldsymbol{u}_j^i \in \mathbb{R}^{n_i \times 3}$. $n_i$ is dimension of data discretisation on geometries $\Omega_i$. Since S2NO is discretisation-invariant, network architectures and parameters can be shared across different geometries $\Omega_i$ and their Laplacian eigenfunctions $\boldsymbol{\Phi}_{\Omega_i}$. With the datasets $\mathcal{D}_i (i=1, \ldots, M)$, we aim to learning a mapping model $\mathcal{G}_{\boldsymbol{\theta}}: \boldsymbol{a}^1 \cup \boldsymbol{a}^2 \cup \ldots \cup \boldsymbol{a}^M \mapsto \boldsymbol{u}^1 \cup \boldsymbol{u}^2 \cup \ldots \cup \boldsymbol{u}^M$. The learning process is mathematically formulated as

\begin{equation}
	\boldsymbol{\theta}^* \in \underset{\boldsymbol{\theta}}{\arg \min } \frac{1}{M} \sum_{i=1}^M \frac{1}{N_i} \sum_{j=1}^{N_i} \frac{\left\|\boldsymbol{u}_j^i-\mathcal{G}_{\boldsymbol{\theta}}\left(\boldsymbol{a}_j^i, \boldsymbol{\Phi}_{\Omega_i}\right)\right\|_2}{\left\|\boldsymbol{u}_j^i\right\|_2} .
\end{equation}

\subsection{S2NO-based inverse design}

The trained S2NO model $\mathcal{G}_{\boldsymbol{\theta}}$ can rapidly and accurately predict the actuated shape of shape-morphing soft material for any given material distribution. By combining an optimisation algorithm with the S2NO model, we can conduct an exhaustive exploration across the design space, thereby enabling efficient inverse engineering. Consider a shape-morphing structure with $K$ designable material voxels, where each voxel has q material types to choose from. Our goal is to find a material distribution from a vast design space consisting of $q^K$ combinations to minimise the error between the achieved shape and the target shape. We use the mean point-to-point distance between two shapes as the objective function for optimisation algorithms. The optimisation process is mathematically formulated as

\begin{equation}
	\boldsymbol{\omega}^* \in \underset{\boldsymbol{\omega}}{\arg \min } \frac{1}{n} \sum_{i=1}^n\left\|\boldsymbol{u}_i^t-\boldsymbol{u}_i^p\right\|_2, \boldsymbol{u}^p=\mathcal{G}_{\boldsymbol{\theta}}(\boldsymbol{a}(\boldsymbol{\omega})),
\end{equation}

where $\boldsymbol{\omega} \in \mathbb{R}^K$ is a vector of design variables, each of which represents the material type for a voxel. $\boldsymbol{u}^t \in \mathbb{R}^{n \times 3}$ is the coordinate field of the discretisation points on the target shape. $\boldsymbol{u}^p \in \mathbb{R}^{n \times 3}$ is the S2NO-predicted shape coordinate field for the material distribution $\boldsymbol{a}(\boldsymbol{\omega})$ corresponding to a solution $\boldsymbol{\omega}$. When the target shape is defined as a surface, the objective function is computed via the target surface and the mid-surface of the predicted shape. This work utilises a population-based genetic algorithm (GA) to explore a vast design space and find an optimal solution for a given target shape. The GA population size is set to 1000 , and the generation size is set to 100 , with crossover and mutation probabilities of 0.75 and 0.2 , respectively. Furthermore, Supplementary Fig. 16 presents a comprehensive process for the multi-resolution optimisation strategy.

\section{Data availability}

Source data are provided with this paper. The data supporting the findings of this work have been deposited in the Google Cloud Drive under accession \url{https://drive.google.com/drive/folders/1OUKJacsNhWbBd7ByzSLCTpTYepj2V0ju?usp=drive_link}. The trained S2NO models and the optimisation results are available at \url{https://github.com/code-cl/S2NO_for_Morphing_Design}.

\section{Code availability}

The codes for model training and optimisation design are available via GitHub at \url{https://github.com/code-cl/S2NO_for_Morphing_Design}.

\section{Acknowledgements}

This work was supported by the National Key R\&D Program of China (No. 2024YFB3310600), the General Program of the National Natural Science Foundation of China (No. 52275491), the Major Program of the National Natural Science Foundation of China (No. 52090052), and the New Cornerstone Science Foundation through the XPLORER PRIZE. This work is partially supported by High Performance Computing Platform of Nanjing University of Aeronautics and Astronautics.

\section{Author contributions statement}

L.C., G.C., X. Liu and Y.L. conceived the research and designed the methodology. L.C., J.S., and X. Lyu implemented the algorithms. J.S. and X. Lyu performed the FE simulations. L.C., J.S., and X. Lyu conducted the experiments and analysed the results. L.W. and Y.L. guided methods and experimental design. G.C., X. Liu, L.W., and Y.L. contributed to the results analysis. Y.L. supervised the project and contributed to securing funding. L.C., G.C., and X. Liu wrote the original draft. All authors contributed to discussions and manuscript preparation.

\section{Competing interests statement}

The authors declare no competing interests.

\bibliographystyle{unsrtnat}
\bibliography{references}  

@article{wang2024performance,
  title={Performance metrics for shape-morphing devices},
  author={Wang, Jue and Chortos, Alex},
  journal={Nature Reviews Materials},
  volume={9},
  number={10},
  pages={738--751},
  year={2024},
  publisher={Nature Publishing Group UK London},
  doi={10.1038/s41578-024-00714-w}
}

@article{he2024programmable,
  title={Programmable responsive metamaterials for mechanical computing and robotics},
  author={He, Qiguang and Ferracin, Samuele and Raney, Jordan R.},
  journal={Nature Computational Science},
  volume={4},
  number={8},
  pages={567--573},
  year={2024},
  publisher={Nature Publishing Group US New York},
  doi={10.1038/s43588-024-00673-w}
}

@article{hu2018small,
  title={Small-scale soft-bodied robot with multimodal locomotion},
  author={Hu, Wenqi and Lum, Guo Zhan and Mastrangeli, Massimo and Sitti, Metin},
  journal={Nature},
  volume={554},
  number={7690},
  pages={81--85},
  year={2018},
  publisher={Nature Publishing Group UK London},
  doi={10.1038/nature25443}
}

@article{bao2025real,
  title={Real-time in situ magnetization reprogramming for soft robotics},
  author={Bao, Xianqiang and Wang, Fan and Zhang, Jianhua and Li, Mingtong and Zhang, Shuaizhong and Ren, Ziyu and Liao, Jiahe and Yan, Yingbo and Kang, Wenbin and Zhang, Rongjing and others},
  journal={Nature},
  volume={645},
  number={8080},
  pages={375--384},
  year={2025},
  publisher={Nature Publishing Group UK London},
  doi={10.1038/s41586-025-09459-0}
}

@article{xu2025transforming,
  title={Transforming machines capable of continuous 3D shape morphing and locking},
  author={Xu, Shiwei and Hu, Xiaonan and Yang, Ruoxi and Zang, Chuanqi and Li, Lei and Xiao, Yue and Liu, Wenbo and Tian, Bocheng and Pang, Wenbo and Bo, Renheng and others},
  journal={Nature Machine Intelligence},
  pages={1--13},
  year={2025},
  publisher={Nature Publishing Group UK London},
  doi={10.1038/s42256-025-01028-4}
}

@article{li2022soft,
  title={Soft actuators for real-world applications},
  author={Li, Meng and Pal, Aniket and Aghakhani, Amirreza and Pena-Francesch, Abdon and Sitti, Metin},
  journal={Nature Reviews Materials},
  volume={7},
  number={3},
  pages={235--249},
  year={2022},
  publisher={Nature Publishing Group UK London},
  doi={10.1038/s41578-021-00389-7}
}

@article{mao2024magnetic,
  title={Magnetic steering continuum robot for transluminal procedures with programmable shape and functionalities},
  author={Mao, Liyang and Yang, Peng and Tian, Chenyao and Shen, Xingjian and Wang, Feihao and Zhang, Hao and Meng, Xianghe and Xie, Hui},
  journal={Nature communications},
  volume={15},
  number={1},
  pages={3759},
  year={2024},
  publisher={Nature Publishing Group UK London},
  doi={10.1038/s41467-024-48058-x}
}

@article{liu2014shape,
  title={Shape memory polymers and their composites in aerospace applications: a review},
  author={Liu, Yanju and Du, Haiyang and Liu, Liwu and Leng, Jinsong},
  journal={Smart materials and structures},
  volume={23},
  number={2},
  pages={023001},
  year={2014},
  publisher={IOP Publishing},
  doi={10.1088/0964-1726/23/2/023001}
}

@article{mahmood2023revolutionizing,
  title={Revolutionizing manufacturing: a review of 4D printing materials, stimuli, and cutting-edge applications},
  author={Mahmood, Ayyaz and Akram, Tehmina and Shenggui, Chen and Chen, Huafu},
  journal={Composites Part B: Engineering},
  volume={266},
  pages={110952},
  year={2023},
  publisher={Elsevier},
  doi={10.1016/j.compositesb.2023.110952}
}

@article{meng2024programmable,
  title={Programmable spatial magnetization stereolithographic printing of biomimetic soft machines with thin-walled structures},
  author={Meng, Xianghe and Li, Shishi and Shen, Xingjian and Tian, Chenyao and Mao, Liyang and Xie, Hui},
  journal={Nature Communications},
  volume={15},
  number={1},
  pages={10442},
  year={2024},
  publisher={Nature Publishing Group UK London},
  doi={10.1038/s41467-024-54773-2}
}

@article{yang2025active,
  title={Active twisting for adaptive droplet collection},
  author={Yang, Yifan and Dai, Zhijun and Chen, Yuzhen and Xu, Fan},
  journal={Nature Computational Science},
  pages={1--9},
  year={2025},
  publisher={Nature Publishing Group US New York},
  doi={10.1038/s43588-025-00786-w}
}

@article{nojoomi2018bioinspired,
  title={Bioinspired 3D structures with programmable morphologies and motions},
  author={Nojoomi, Amirali and Arslan, Hakan and Lee, Kwan and Yum, Kyungsuk},
  journal={Nature communications},
  volume={9},
  number={1},
  pages={3705},
  year={2018},
  publisher={Nature Publishing Group UK London},
  doi={10.1038/s41467-018-05569-8}
}

@article{gladman2016biomimetic,
  title={Biomimetic 4D printing},
  author={Sydney Gladman, A. and Matsumoto, Elisabetta A. and Nuzzo, Ralph G. and Mahadevan, L. and Lewis, Jennifer A.},
  journal={Nature materials},
  volume={15},
  number={4},
  pages={413--418},
  year={2016},
  publisher={Nature Publishing Group UK London},
  doi={10.1038/nmat4544}
}

@article{kim2018printing,
  title={Printing ferromagnetic domains for untethered fast-transforming soft materials},
  author={Kim, Yoonho and Yuk, Hyunwoo and Zhao, Ruike and Chester, Shawn A. and Zhao, Xuanhe},
  journal={Nature},
  volume={558},
  number={7709},
  pages={274--279},
  year={2018},
  publisher={Nature Publishing Group UK London},
  doi={10.1038/s41586-018-0185-0}
}

@article{wang2023programmable,
  title={Programmable spatial deformation by controllable off-center freestanding 4D printing of continuous fiber reinforced liquid crystal elastomer composites},
  author={Wang, Qingrui and Tian, Xiaoyong and Zhang, Daokang and Zhou, Yanli and Yan, Wanquan and Li, Dichen},
  journal={Nature Communications},
  volume={14},
  number={1},
  pages={3869},
  year={2023},
  publisher={Nature Publishing Group UK London},
  doi={10.1038/s41467-023-39566-3}
}

@article{sun2025stimuli,
  title={Stimuli-responsive shape-morphing soft actuators: metrics, materials, mechanism, design and applications},
  author={Sun, Linchao and Li, Zhong and Zhang, Yan and Lu, Yao and Zhang, Shiguo},
  journal={Progress in Materials Science},
  volume={155},
  pages={101531},
  year={2026},
  publisher={Elsevier},
  doi={10.1016/j.pmatsci.2025.101531}
}

@article{yarali2024d,
  title={4D printing for biomedical applications},
  author={Yarali, Ebrahim and Mirzaali, Mohammad J. and Ghalayaniesfahani, Ava and Accardo, Angelo and Diaz-Payno, Pedro J. and Zadpoor, Amir A.},
  journal={Advanced Materials},
  volume={36},
  number={31},
  pages={2402301},
  year={2024},
  publisher={Wiley Online Library},
  doi={10.1002/adma.202402301}
}

@article{wang2021evolutionary,
  title={Evolutionary design of magnetic soft continuum robots},
  author={Wang, Liu and Zheng, Dongchang and Harker, Pablo and Patel, Aman B. and Guo, Chuan Fei and Zhao, Xuanhe},
  journal={Proceedings of the National Academy of Sciences},
  volume={118},
  number={21},
  pages={e2021922118},
  year={2021},
  publisher={National Academy of Sciences},
  doi={10.1073/pnas.2021922118}
}

@article{peng2023controllable,
  title={Controllable deformation design for 4D-printed active composite structure: optimization, simulation, and experimental verification},
  author={Peng, Xiang and Liu, Guoao and Wang, Jun and Li, Jiquan and Wu, Huaping and Jiang, Shaofei and Yi, Bing},
  journal={Composites Science and Technology},
  volume={243},
  pages={110265},
  year={2023},
  publisher={Elsevier},
  doi={10.1016/j.compscitech.2023.110265}
}

@article{averitt2025artificial,
  title={Artificial Intelligence and Computing for Active Metamaterial Design: A Perspective},
  author={Averitt, Samantha and Sim, Jay and Zhao, Ruike Renee},
  journal={Journal of Applied Mechanics},
  volume={92},
  number={9},
  pages={094001},
  year={2025},
  publisher={American Society of Mechanical Engineers},
  doi={10.1115/1.4068647}
}

@article{nojoomi20212d,
  title={2D material programming for 3D shaping},
  author={Nojoomi, Amirali and Jeon, Junha and Yum, Kyungsuk},
  journal={Nature communications},
  volume={12},
  number={1},
  pages={603},
  year={2021},
  publisher={Nature Publishing Group UK London},
  doi={10.1038/s41467-021-20934-w}
}

@article{xia2025inverse,
  title={Inverse programming of ferromagnetic domains for 3D curved surfaces of soft materials},
  author={Xia, Neng and Jin, Dongdong and Yang, Zhengxin and Pan, Chengfeng and Su, Lin and Zhang, Moqiu and Wang, Xin and Xu, Zirong and Guo, Zichang and Pan, Longyu and others},
  journal={Nature Synthesis},
  pages={1--13},
  year={2025},
  publisher={Nature Publishing Group UK London},
  doi={10.1038/s44160-025-00746-2}
}

@article{sun2024machine,
 title={Machine learning-enabled forward prediction and inverse design of 4D-printed active plates},
  author={Sun, Xiaohao and Yue, Liang and Yu, Luxia and Forte, Connor T. and Armstrong, Connor D. and Zhou, Kun and Demoly, Fr{\'e}d{\'e}ric and Zhao, Ruike Renee and Qi, H. Jerry},
  journal={Nature Communications},
  volume={15},
  number={1},
  pages={5509},
  year={2024},
  publisher={Nature Publishing Group UK London},
  doi={10.1038/s41467-024-49775-z}
}

@article{li2023general,
  title={A general theoretical scheme for shape-programming of incompressible hyperelastic shells through differential growth},
  author={Li, Zhanfeng and Wang, Jiong and Hossain, Mokarram and Kadapa, Chennakesava},
  journal={International Journal of Solids and Structures},
  volume={265},
  pages={112128},
  year={2023},
  publisher={Elsevier},
  doi={10.1016/j.ijsolstr.2023.112128}
}

@article{raabe2023accelerating,
  title={Accelerating the design of compositionally complex materials via physics-informed artificial intelligence},
  author={Raabe, Dierk and Mianroodi, Jaber Rezaei and Neugebauer, J{\"o}rg},
  journal={Nature computational science},
  volume={3},
  number={3},
  pages={198--209},
  year={2023},
  publisher={Nature Publishing Group US New York},
  doi={10.1038/s43588-023-00412-7}
}

@article{sun2024perspective,
  title={Perspective: Machine learning in design for 3D/4D printing},
  author={Sun, Xiaohao and Zhou, Kun and Demoly, Fr{\'e}d{\'e}ric and Zhao, Ruike Renee and Qi, H. Jerry},
  journal={Journal of Applied Mechanics},
  volume={91},
  number={3},
  pages={030801},
  year={2024},
  publisher={American Society of Mechanical Engineers},
  doi={10.1115/1.4063684}
}

@article{cheng2023programming,
  title={Programming 3D curved mesosurfaces using microlattice designs},
  author={Cheng, Xu and Fan, Zhichao and Yao, Shenglian and Jin, Tianqi and Lv, Zengyao and Lan, Yu and Bo, Renheng and Chen, Yitong and Zhang, Fan and Shen, Zhangming and others},
  journal={Science},
  volume={379},
  number={6638},
  pages={1225--1232},
  year={2023},
  publisher={American Association for the Advancement of Science},
  doi={10.1126/science.adf3824}
}

@article{forte2022inverse,
  title={Inverse design of inflatable soft membranes through machine learning},
  author={Forte, Antonio Elia and Hanakata, Paul Z. and Jin, Lishuai and Zari, Emilia and Zareei, Ahmad and Fernandes, Matheus C. and Sumner, Laura and Alvarez, Jonathan and Bertoldi, Katia},
  journal={Advanced Functional Materials},
  volume={32},
  number={16},
  pages={2111610},
  year={2022},
  publisher={Wiley Online Library},
  doi={10.1002/adfm.202111610}
}

@article{karacakol2025data,
  title={Data-driven design of shape-programmable magnetic soft materials},
  author={Karacakol, Alp C. and Alapan, Yunus and Demir, Sinan O. and Sitti, Metin},
  journal={Nature Communications},
  volume={16},
  number={1},
  pages={2946},
  year={2025},
  publisher={Nature Publishing Group UK London},
  doi={10.1038/s41467-025-58091-z}
}

@article{sun2022machine,
  title={Machine learning-evolutionary algorithm enabled design for 4D-printed active composite structures},
  author={Sun, Xiaohao and Yue, Liang and Yu, Luxia and Shao, Han and Peng, Xirui and Zhou, Kun and Demoly, Fr{\'e}d{\'e}ric and Zhao, Ruike and Qi, H. Jerry},
  journal={Advanced Functional Materials},
  volume={32},
  number={10},
  pages={2109805},
  year={2022},
  publisher={Wiley Online Library},
  doi={10.1002/adfm.202109805}
}

@article{wang2023scientific,
  title={Scientific discovery in the age of artificial intelligence},
  author={Wang, Hanchen and Fu, Tianfan and Du, Yuanqi and Gao, Wenhao and Huang, Kexin and Liu, Ziming and Chandak, Payal and Liu, Shengchao and Van Katwyk, Peter and Deac, Andreea and others},
  journal={Nature},
  volume={620},
  number={7972},
  pages={47--60},
  year={2023},
  publisher={Nature Publishing Group UK London},
  doi={10.1038/s41586-023-06221-2}
}

@article{azizzadenesheli2024neural,
  title={Neural operators for accelerating scientific simulations and design},
  author={Azizzadenesheli, Kamyar and Kovachki, Nikola and Li, Zongyi and Liu-Schiaffini, Miguel and Kossaifi, Jean and Anandkumar, Anima},
  journal={Nature Reviews Physics},
  volume={6},
  number={5},
  pages={320--328},
  year={2024},
  publisher={Nature Publishing Group UK London},
  doi={10.1038/s42254-024-00712-5}
}

@article{xia2022responsive,
  title={Responsive materials architected in space and time},
  author={Xia, Xiaoxing and Spadaccini, Christopher M. and Greer, Julia R.},
  journal={Nature Reviews Materials},
  volume={7},
  number={9},
  pages={683--701},
  year={2022},
  publisher={Nature Publishing Group UK London},
  doi={10.1038/s41578-022-00450-z}
}

@article{liu20234d,
  title={4D printing of mechanically robust PLA/TPU/Fe3O4 magneto-responsive shape memory polymers for smart structures},
  author={Liu, Han and Wang, Feifan and Wu, Wenyang and Dong, Xufeng and Sang, Lin},
  journal={Composites Part B: Engineering},
  volume={248},
  pages={110382},
  year={2023},
  publisher={Elsevier},
  doi={10.1016/j.compositesb.2022.110382}
}

@article{lanthaler2023operator,
  title={Operator learning with PCA-Net: upper and lower complexity bounds},
  author={Lanthaler, Samuel},
  journal={Journal of Machine Learning Research},
  volume={24},
  number={318},
  pages={1--67},
  year={2023},
  publisher={Microtome Publishing},
  url={http://jmlr.org/papers/v24/23-0478.html}
}

@article{bhattacharya2021model,
  title={Model reduction and neural networks for parametric PDEs},
  author={Bhattacharya, Kaushik and Hosseini, Bamdad and Kovachki, Nikola B. and Stuart, Andrew M.},
  journal={The SMAI journal of computational mathematics},
  volume={7},
  pages={121--157},
  year={2021},
  doi={10.5802/smai-jcm.74}
}

@article{lu2021learning,
  title={Learning nonlinear operators via DeepONet based on the universal approximation theorem of operators},
  author={Lu, Lu and Jin, Pengzhan and Pang, Guofei and Zhang, Zhongqiang and Karniadakis, George Em},
  journal={Nature machine intelligence},
  volume={3},
  number={3},
  pages={218--229},
  year={2021},
  publisher={Nature Publishing Group UK London},
  doi={10.1038/s42256-021-00302-5}
}

@article{lu2022comprehensive,
  title={A comprehensive and fair comparison of two neural operators (with practical extensions) based on fair data},
  author={Lu, Lu and Meng, Xuhui and Cai, Shengze and Mao, Zhiping and Goswami, Somdatta and Zhang, Zhongqiang and Karniadakis, George Em},
  journal={Computer Methods in Applied Mechanics and Engineering},
  volume={393},
  pages={114778},
  year={2022},
  publisher={Elsevier},
  doi={10.1016/j.cma.2022.114778}
}

@article{chen2024learning,
 title={Learning neural operators on riemannian manifolds},
  author={Chen, Gengxiang and Liu, Xu and Meng, Qinglu and Chen, Lu and Liu, Changqing and Li, Yingguang},
  journal={National Science Open},
  volume={3},
  number={6},
  pages={20240001},
  year={2024},
  publisher={China Science Publishing \& Media Ltd. and EDP Sciences},
  doi={10.1360/nso/20240001}
}

@inproceedings{wu2024transolver,
  title={Transolver: A fast transformer solver for pdes on general geometries},
  author={Wu, Haixu and Luo, Huakun and Wang, Haowen and Wang, Jianmin and Long, Mingsheng},
  booktitle={Proceedings of the 41st International Conference on Machine Learning},
  pages={53681--53705},
  year={2024},
  volume={235},
  series={Proceedings of Machine Learning Research},
  publisher={PMLR},
  url={https://proceedings.mlr.press/v235/wu24r.html}
}

@article{kovachki2023neural,
  title={Neural operator: Learning maps between function spaces with applications to pdes},
  author={Kovachki, Nikola and Li, Zongyi and Liu, Burigede and Azizzadenesheli, Kamyar and Bhattacharya, Kaushik and Stuart, Andrew and Anandkumar, Anima},
  journal={Journal of Machine Learning Research},
  volume={24},
  number={89},
  pages={1--97},
  year={2023},
  url={http://jmlr.org/papers/v24/21-1524.html}
}

@article{li2020fourier,
  title={Fourier neural operator for parametric partial differential equations},
  author={Li, Zongyi and Kovachki, Nikola and Azizzadenesheli, Kamyar and Liu, Burigede and Bhattacharya, Kaushik and Stuart, Andrew and Anandkumar, Anima},
  journal={arXiv preprint arXiv:2010.08895},
  year={2020},
  url={https://arxiv.org/abs/2010.08895}
}

@article{reuter2006laplace,
 title={Laplace--Beltrami spectra as ‘Shape-DNA’of surfaces and solids},
  author={Reuter, Martin and Wolter, Franz-Erich and Peinecke, Niklas},
  journal={Computer-Aided Design},
  volume={38},
  number={4},
  pages={342--366},
  year={2006},
  publisher={Elsevier},
  doi={10.1016/j.cad.2005.10.011}
}

@article{vaswani2017attention,
  title={Attention is all you need},
  author={Vaswani, Ashish and Shazeer, Noam and Parmar, Niki and Uszkoreit, Jakob and Jones, Llion and Gomez, Aidan N and Kaiser, {\L}ukasz and Polosukhin, Illia},
  journal={Advances in neural information processing systems},
  volume={30},
  year={2017},
  url = {https://proceedings.neurips.cc/paper_files/paper/2017/file/3f5ee243547dee91fbd053c1c4a845aa-Paper.pdf},
}

@inproceedings{su2024multiscale,
  title={Multiscale attention wavelet neural operator for capturing steep trajectories in biochemical systems},
  author={Su, Jiayang and Ma, Junbo and Tong, Songyang and Xu, Enze and Chen, Minghan},
  booktitle={Proceedings of the AAAI Conference on Artificial Intelligence},
  volume={38},
  number={13},
  pages={15100--15107},
  year={2024},
  doi={10.1609/aaai.v38i13.29432}
}






\end{document}


\maketitle
\vspace{8em} 
\tableofcontents
\newpage

\section{Supplementary Notes}

\subsection{Details for the S2NO model}

The overall framework of the S2NO model is illustrated in Fig. 1d-f, which consists of a lifting layer $P$ to increase the channel dimension of the data, multiple S2NO layers to learn the operator, and a projecting layer $Q$ to project the data back to the original output channel dimension. The lifting layer $P$ and projecting layer $Q$ can be implemented through two pointwise neural networks. Each S2NO layer comprises two key components: spectral convolution and spatial convolution, which are designed to capture global and local morphing behaviours Moreover, residual connections, layer normalisation, and feed forward, which are analogous to those in Transformer models \cite{vaswani2017attention}, are introduced to enhance generalisation performance \cite{su2024multiscale}.

\textbf{Spectral convolution.} The spectral convolution is a block consisting of an encoder, approximator, and decoder (see Fig. 1e). Denote the input data for the spectral convolution block as $\boldsymbol{v}_x \in \mathbb{R}^{n \times d_c}$, where $n$ is the spatial dimension of data, $d_c$ is the channel dimension after lifting. The encoder $\mathcal{E}$, similar to the fast Fourier transform, transforms the highdimensional field data $\boldsymbol{v}_x \in \mathbb{R}^{n \times d_c}$ to the low-dimensional frequency domain $\varepsilon\left(\boldsymbol{v}_x\right) \in \mathbb{R}^{k \times d_c}$ based on spectral decomposition using Laplacian eigenfunctions of the complex geometry, where $k$ is the number of Laplacian eigenfunctions. The approximator $\mathcal{R}$ is a learnable mapping in the frequency domain, and $\mathcal{R} \cdot\left(\varepsilon\left(\boldsymbol{v}_x\right)\right) \in \mathbb{R}^{k \times d_c}$. The decoder $\mathcal{D}$, similar to the inverse Fourier transform, reconstructs the coefficient data to the original spatial dimension based on spectral reconstruction using Laplacian eigenfunctions, and $\mathcal{D}(\mathcal{R}$. $\left.\left(\varepsilon\left(\boldsymbol{v}_x\right)\right)\right) \in \mathbb{R}^{n \times d_c}$. Here, the approximator is constructed with a multi-head architecture to process features in parallel spectral spaces. Laplacian eigenfunctions of the geometry can be obtained by solving the eigenvalue problem of the Laplacian $\Delta \boldsymbol{\phi}_i=\lambda_i \boldsymbol{\phi}_i$ using the LaPy Python library.

\textbf{Spatial convolution.} Spatial convolution consisting of a graph convolutional layer, an input gate, and an output gate (see Fig. 1f). Graph convolutional layers can capture local features by aggregating information from neighbouring nodes. Two gates are designed with sigmoid activation functions to learn the weights of the contributions of spatial convolution, enabling the judicious fusion of local and global features. The input gate and output gate can be represented as $\prod_i\left(\boldsymbol{h}_i\right)=\sigma\left(\boldsymbol{W}_i \boldsymbol{h}_i+\boldsymbol{b}_i\right)$ and $\prod_o\left(\boldsymbol{h}_o\right)=\sigma\left(\boldsymbol{W}_o \boldsymbol{h}_o+\boldsymbol{b}_o\right)$. The output of this block for the input $\boldsymbol{v}_x$ can be represented as $\Pi_o \odot \operatorname{Graph} \operatorname{Conv}\left(\Pi_i \odot \boldsymbol{v}_x\right)$. The graph convolution operation is implemented using the Torch Geometric Python library. Based on our experimental analysis, global deformation extraction via spectral convolution constitutes the primary contributor to model performance. In contrast, spatial convolution enhances model performance in scenarios involving complex local deformations. In practical implementation, the two modules may be combined according to the specific morphing behaviour.

\subsection{Implementation details for different data-driven models}

In this work, we train data-driven models to predict the actuated shapes of shape-morphing soft materials with complex geometries under different material distributions. Our experiments involve six modelling methods: PODNN, DeepONet, POD-DeepONet, NORM, Transolver, and S2NO. The specific architecture configuration and training parameters for each model are detailed in the Supplementary Table \ref{tab:S1}. The details of each method are as follows.

\textbf{PODNN:} Based on the concept of dimensionality reduction, PODNN transforms the high-dimensional input-output mapping into the low-dimensional mapping \cite{lanthaler2023operator,bhattacharya2021model}. A set of orthogonal bases is obtained by performing proper orthogonal decomposition (POD) on the actuated shape coordinate fields of the training dataset. A neural network (NN) is used to learn the mapping relationship between material voxel vectors and POD coefficients. The reconstructed actuated shape is obtained by combining the coefficients predicted by NN with POD bases.

\textbf{DeepONet:} The Deep Operator Network (DeepONet) comprises a branch net for encoding discrete input functions and a trunk net for encoding the location variables of output functions separately, which are then merged via a dot product to compute the output \cite{lu2021learning}. We train three DeepONet models to predict the coordinates in the x, y, and z directions, respectively. 

\textbf{POD-DeepONet:} POD-DeepONet replaces DeepONet’s trunk net with POD bases derived from the training data to accelerate its training and enhance its accuracy \cite{lu2022comprehensive}. 

\textbf{NORM:} Neural operator on Riemannian manifold (NORM) transforms function-to-function mapping into a finite-dimensional mapping in the Laplacian eigenfunctions’ subspace of geometry, thereby enabling the mapping between functions on complex geometries \cite{chen2024learning}. 

\textbf{Transolver:} Transolver uses a physics-attention to split the discretised domain into a series of learnable slices of flexible shapes, enabling Transolver to learn the intrinsic physical states hidden behind the discretised geometries \cite{wu2024transolver}.

\clearpage

\section{Supplementary Figures}

\begin{figure}[htbp]
	\centering
	\includegraphics[width=1.0\linewidth]{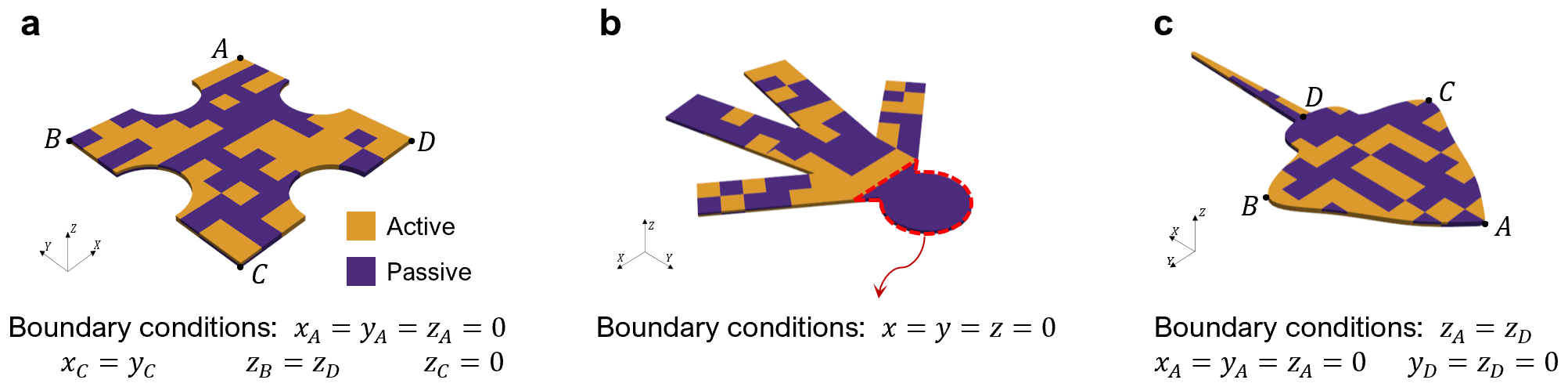}
	\caption{ The shapes, material voxels, and boundary conditions of FE simulations for the dart case (\textbf{a}), human case (\textbf{b}), and stingray case (\textbf{c}). All shapes have a thickness of 1 mm and are subdivided into two material voxels in the thickness direction. The number of material voxels is 152, 129, and 160, respectively.}
	\label{fig:fig1}
\end{figure}

\vspace{8em} 
\begin{figure}[htbp]
	\centering
	\includegraphics[width=1.0\linewidth]{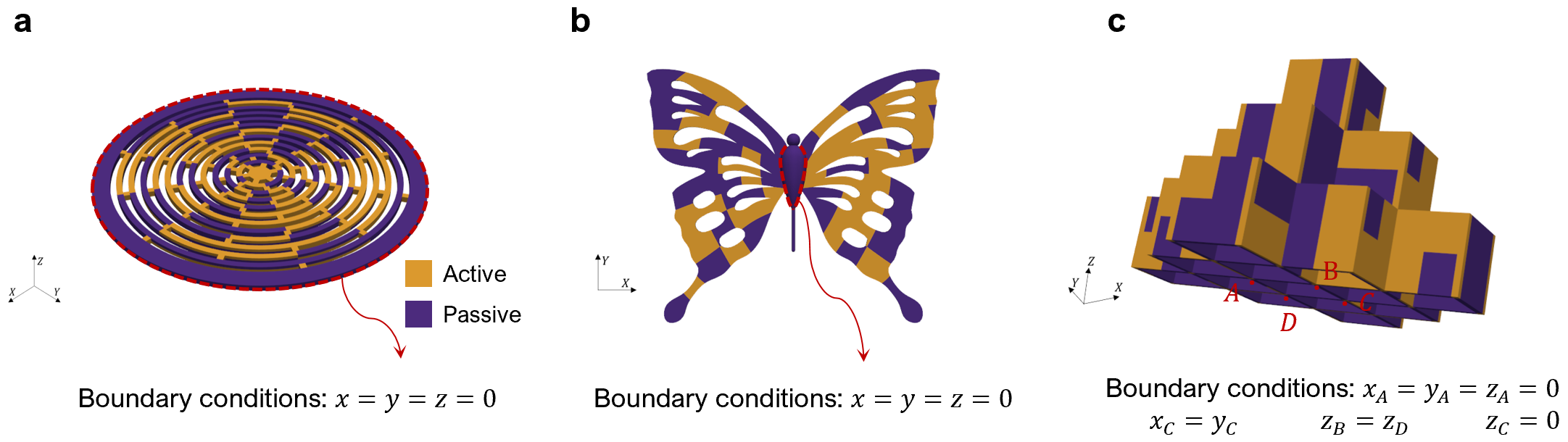}
	\caption{The shapes, material voxels, and boundary conditions of FE simulations for the dome case (\textbf{a}), butterfly case (\textbf{b}), and 3D thin-walled structure case (\textbf{c}). All structures are subdivided into two material voxels in the thickness direction. The number of designable material voxels is 264, 196, and 224, respectively.}
	\label{fig:fig2}
\end{figure}

\begin{figure}[htbp]
	\centering
	\includegraphics[width=1.0\linewidth]{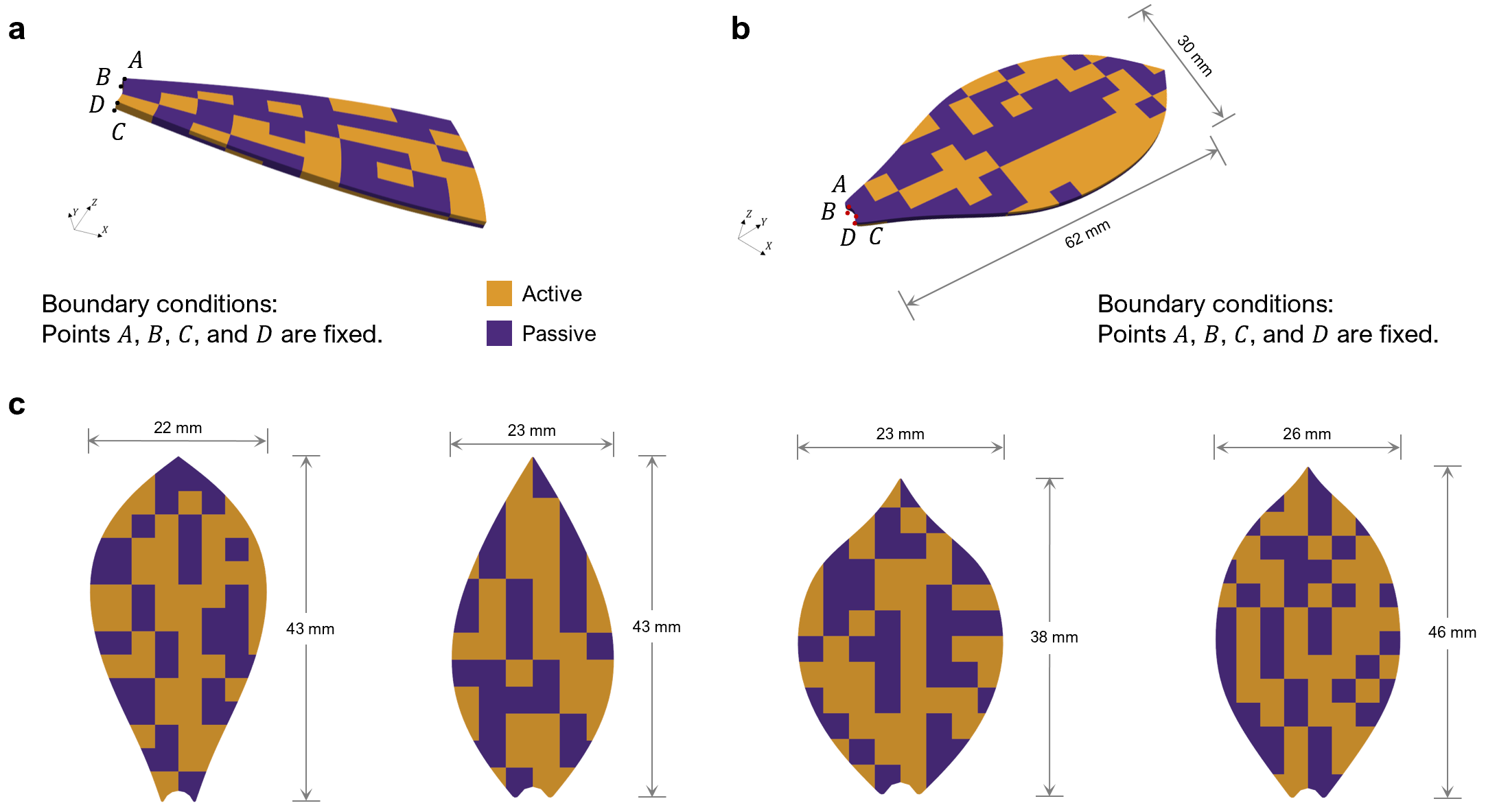}
	\caption{The shapes, material voxels and boundary conditions of FE simulations for the blade case (\textbf{a}) and five petals (\textbf{b}, \textbf{c}). The boundary conditions for the four petals (\textbf{c}) and the single petal (\textbf{b}) are identical. All shapes are subdivided into two material voxels in the thickness direction. The number of material voxels is 120, 180, 160, 120, 152, and 164, respectively.}
	\label{fig:fig3}
\end{figure}

\clearpage
\begin{figure}
	\centering
	\includegraphics[width=1.0\linewidth]{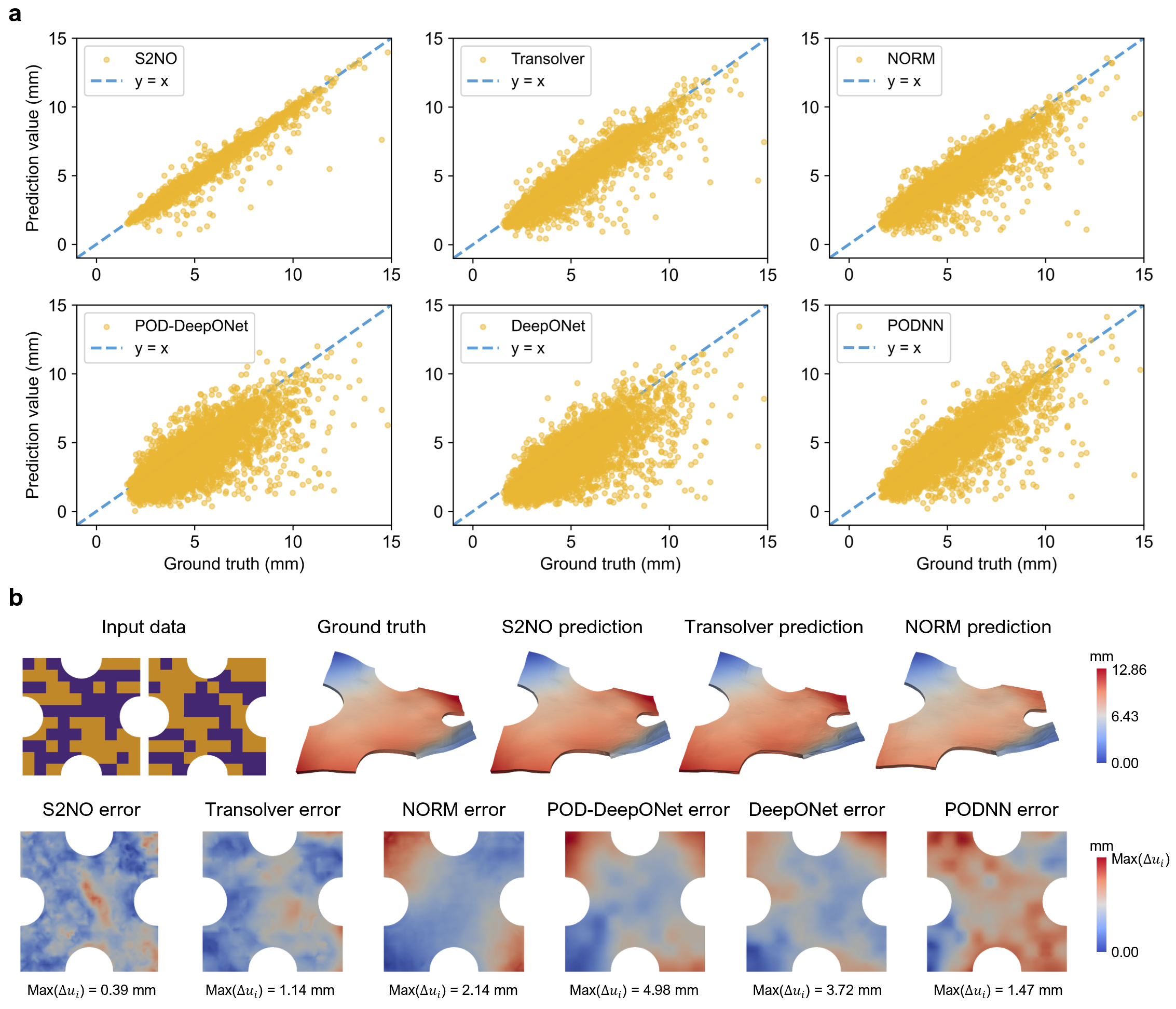}
	\caption{Prediction results of the S2NO model and baseline models for the dart case. \textbf{a}, Scatter plot of prediction values versus ground truth of the maximum deformation for each test sample. Results closer to the reference line $y=x$ indicate higher predictive accuracy of the model. \textbf{b}, Input material distribution, ground truth, prediction results, and error maps for a data sample randomly selected from the test dataset are presented to compare the predictive performance of different models visually. $\operatorname{Max}\left(\Delta u_i\right)$ represents the maximum value of the point-to-point distance between the predicted shape and the ground truth. The $\operatorname{Max}\left(\Delta u_i\right)$ of S2NO is merely 0.39 mm.}
	\label{fig:fig4}
\end{figure}

\clearpage
\begin{figure}
	\centering
	\includegraphics[width=1.0\linewidth]{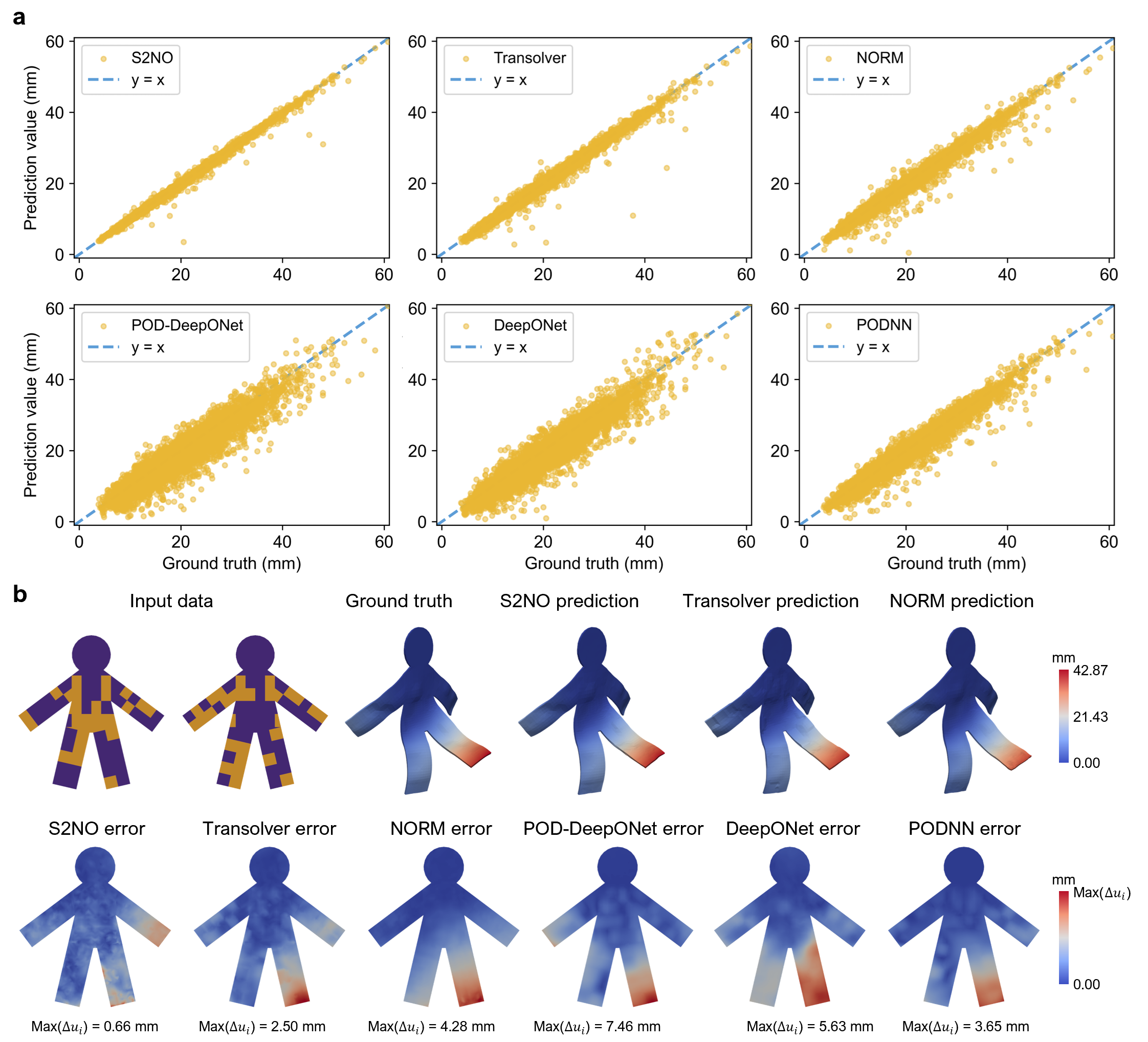}
	\caption{Prediction results of the S2NO model and baseline models for the human case. \textbf{a}, Scatter plot of prediction values versus ground truth of the maximum deformation for each test sample. Results closer to the reference line $y=x$ indicate higher predictive accuracy of the model. \textbf{b}, Input material distribution, ground truth, prediction results, and error maps for a data sample randomly selected from the test dataset are presented to compare the predictive performance of different models visually. $\operatorname{Max}\left(\Delta u_i\right)$ represents the maximum value of the point-to-point distance between the predicted shape and the ground truth. The $\operatorname{Max}\left(\Delta u_i\right)$ of S2NO is merely 0.66 mm.}
	\label{fig:fig5}
\end{figure}

\clearpage
\begin{figure}
	\centering
	\includegraphics[width=1.0\linewidth]{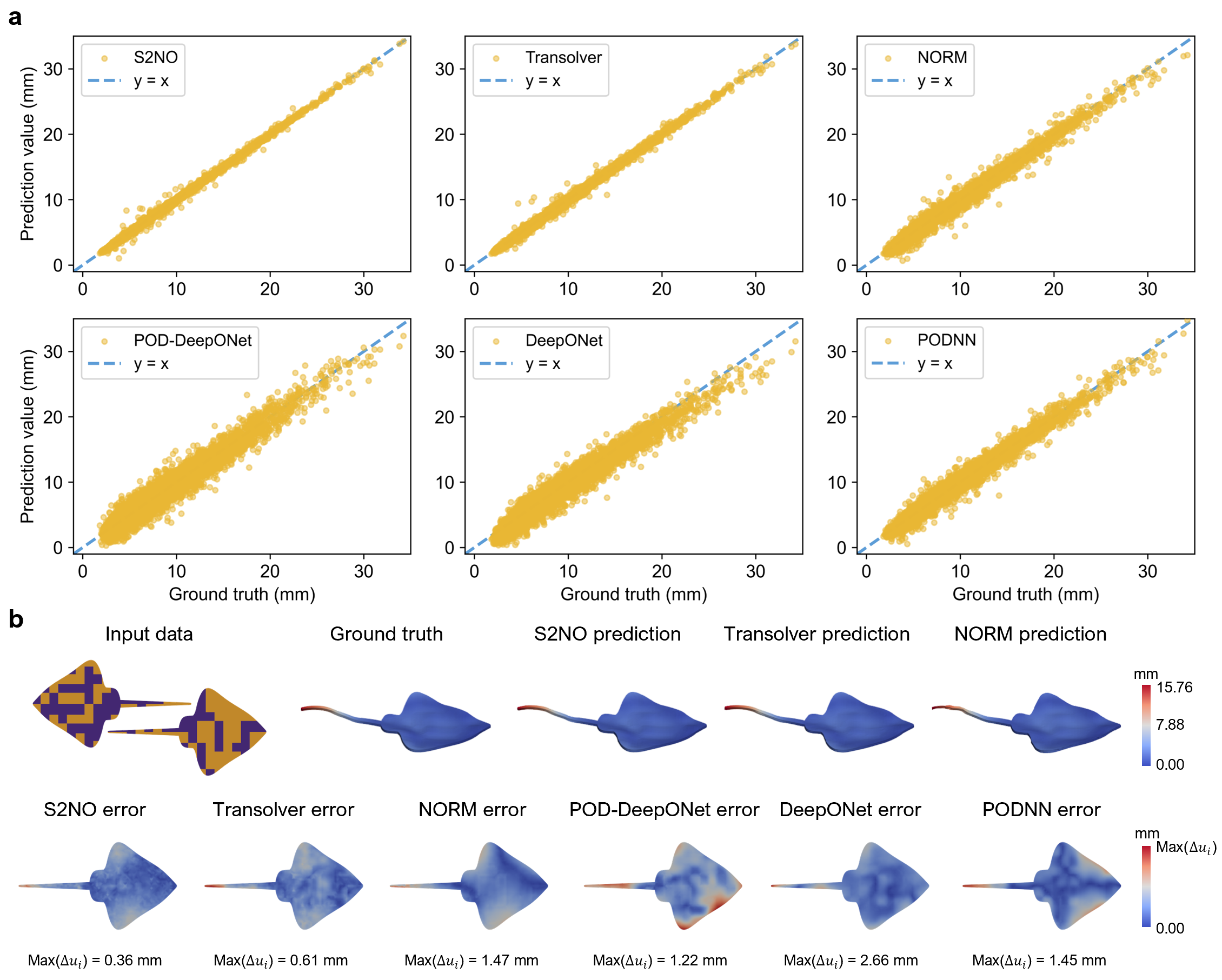}
	\caption{Prediction results of the S2NO model and baseline models for the stingray case. \textbf{a}, Scatter plot of prediction values versus ground truth of the maximum deformation for each test sample. Results closer to the reference line $y=x$ indicate higher predictive accuracy of the model. \textbf{b}, Input material distribution, ground truth, prediction results, and error maps for a data sample randomly selected from the test dataset are presented to compare the predictive performance of different models visually. $\operatorname{Max}\left(\Delta u_i\right)$ represents the maximum value of the point-to-point distance between the predicted shape and the ground truth. The $\operatorname{Max}\left(\Delta u_i\right)$ of S2NO is merely 0.36 mm.}
	\label{fig:fig6}
\end{figure}

\clearpage
\begin{figure}
	\centering
	\includegraphics[width=1.0\linewidth]{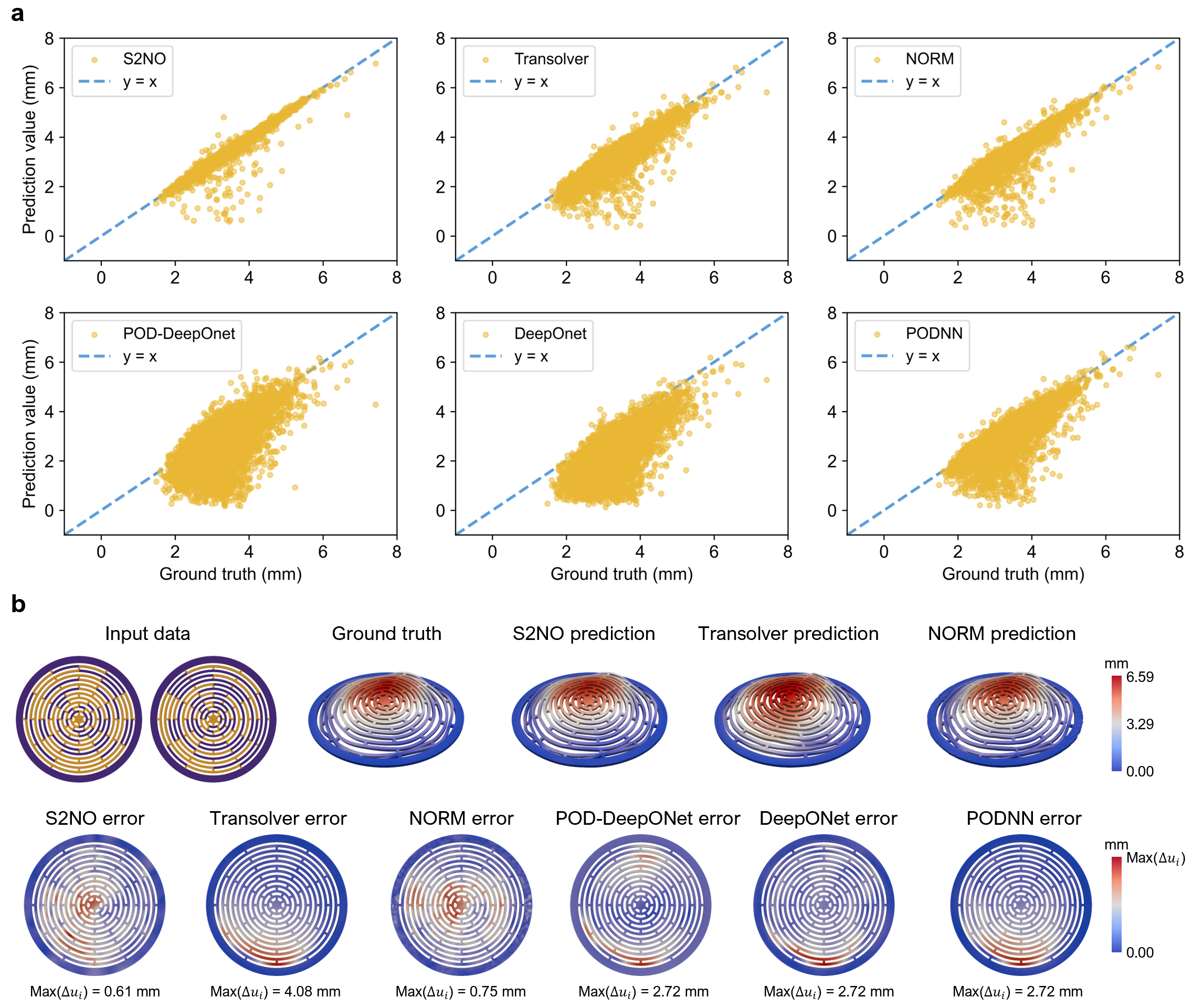}
	\caption{Prediction results of the S2NO model and baseline models for the dome case. \textbf{a}, Scatter plot of prediction values versus ground truth of the maximum deformation for each test sample. Results closer to the reference line $y=x$ indicate higher predictive accuracy of the model. \textbf{b}, Input material distribution, ground truth, prediction results, and error maps for a data sample randomly selected from the test dataset are presented to compare the predictive performance of different models visually. $\operatorname{Max}\left(\Delta u_i\right)$ represents the maximum value of the point-to-point distance between the predicted shape and the ground truth. The $\operatorname{Max}\left(\Delta u_i\right)$ of S2NO is merely 0.61 mm.}
	\label{fig:fig7}
\end{figure}

\clearpage
\begin{figure}
	\centering
	\includegraphics[width=1.0\linewidth]{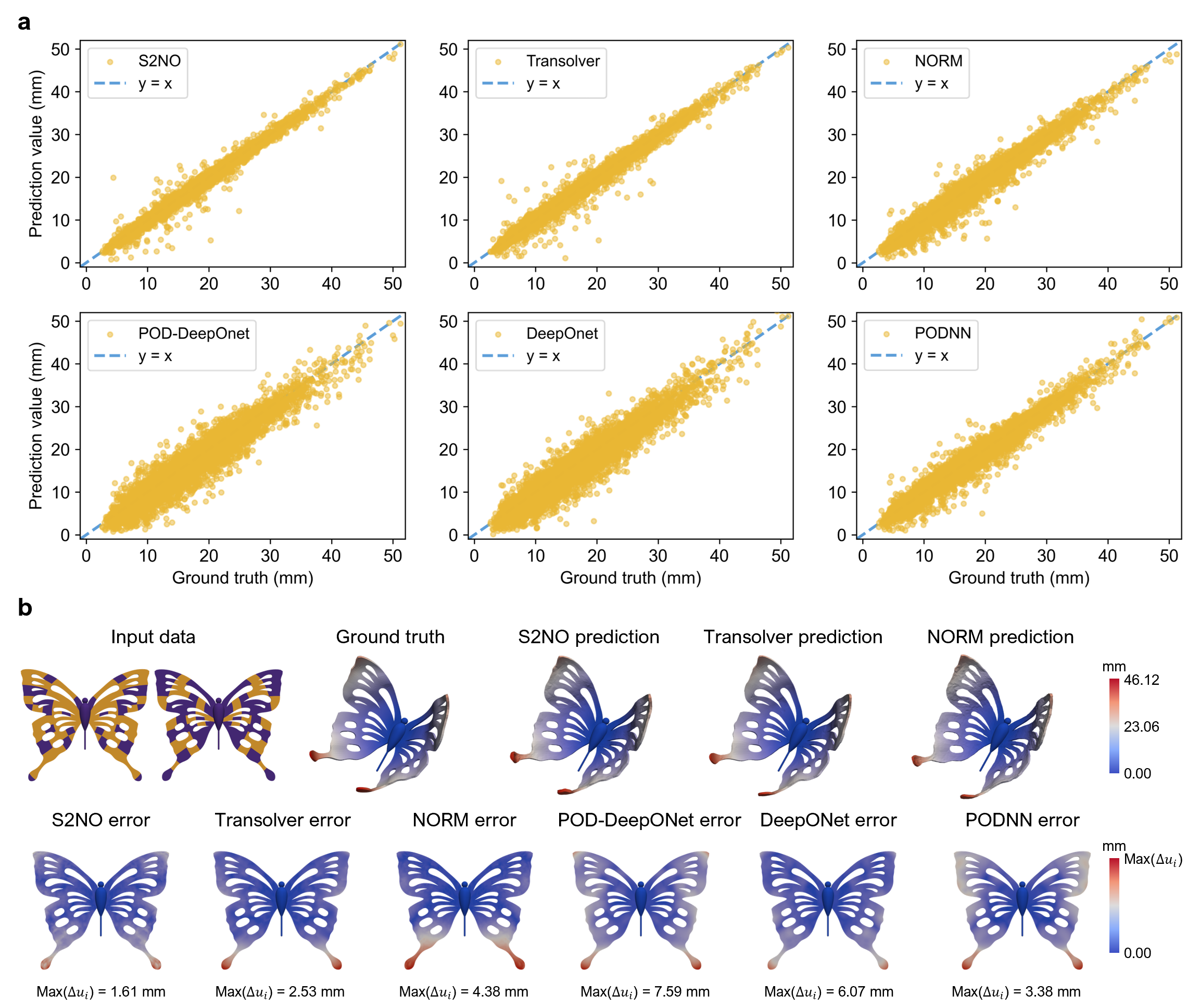}
	\caption{Prediction results of the S2NO model and baseline models for the butterfly case. \textbf{a}, Scatter plot of prediction values versus ground truth of the maximum deformation for each test sample. Results closer to the reference line $y=x$ indicate higher predictive accuracy of the model. \textbf{b}, Input material distribution, ground truth, prediction results, and error maps for a data sample randomly selected from the test dataset are presented to compare the predictive performance of different models visually. $\operatorname{Max}\left(\Delta u_i\right)$ represents the maximum value of the point-to-point distance between the predicted shape and the ground truth. The $\operatorname{Max}\left(\Delta u_i\right)$ of S2NO is merely 1.61 mm.}
	\label{fig:fig8}
\end{figure}

\clearpage
\begin{figure}
	\centering
	\includegraphics[width=1.0\linewidth]{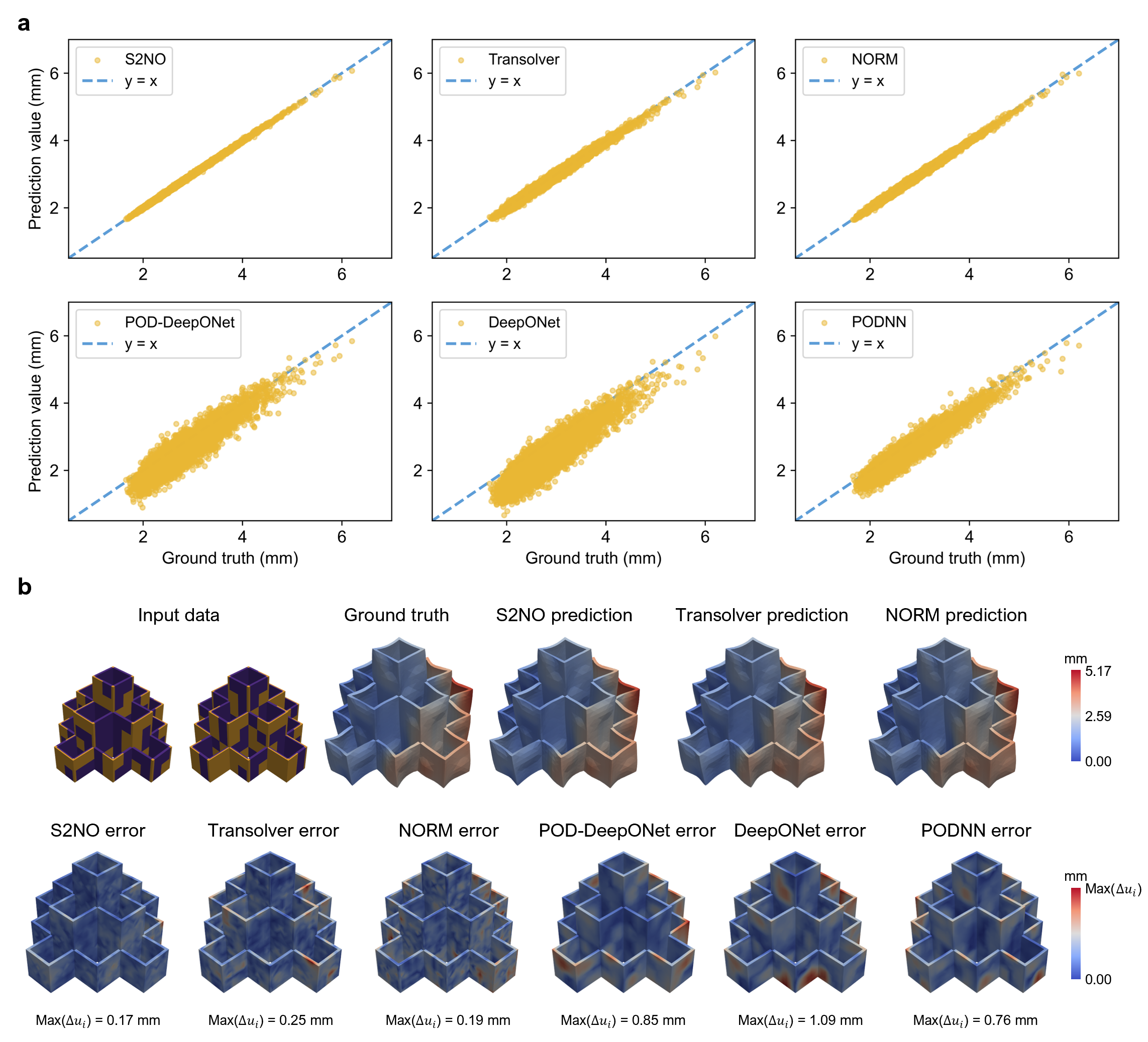}
	\caption{Prediction results of the S2NO model and baseline models for the 3D thin-walled structure case. \textbf{a}, Scatter plot of prediction values versus ground truth of the maximum deformation for each test sample. Results closer to the reference line $y=x$ indicate higher predictive accuracy of the model. \textbf{b}, Input material distribution, ground truth, prediction results, and error maps for a data sample randomly selected from the test dataset are presented to compare the predictive performance of different models visually. $\operatorname{Max}\left(\Delta u_i\right)$ represents the maximum value of the point-to-point distance between the predicted shape and the ground truth. The $\operatorname{Max}\left(\Delta u_i\right)$ of S2NO is merely 0.17 mm.}
	\label{fig:fig9}
\end{figure}

\clearpage
\begin{figure}
	\centering
	\includegraphics[width=1.0\linewidth]{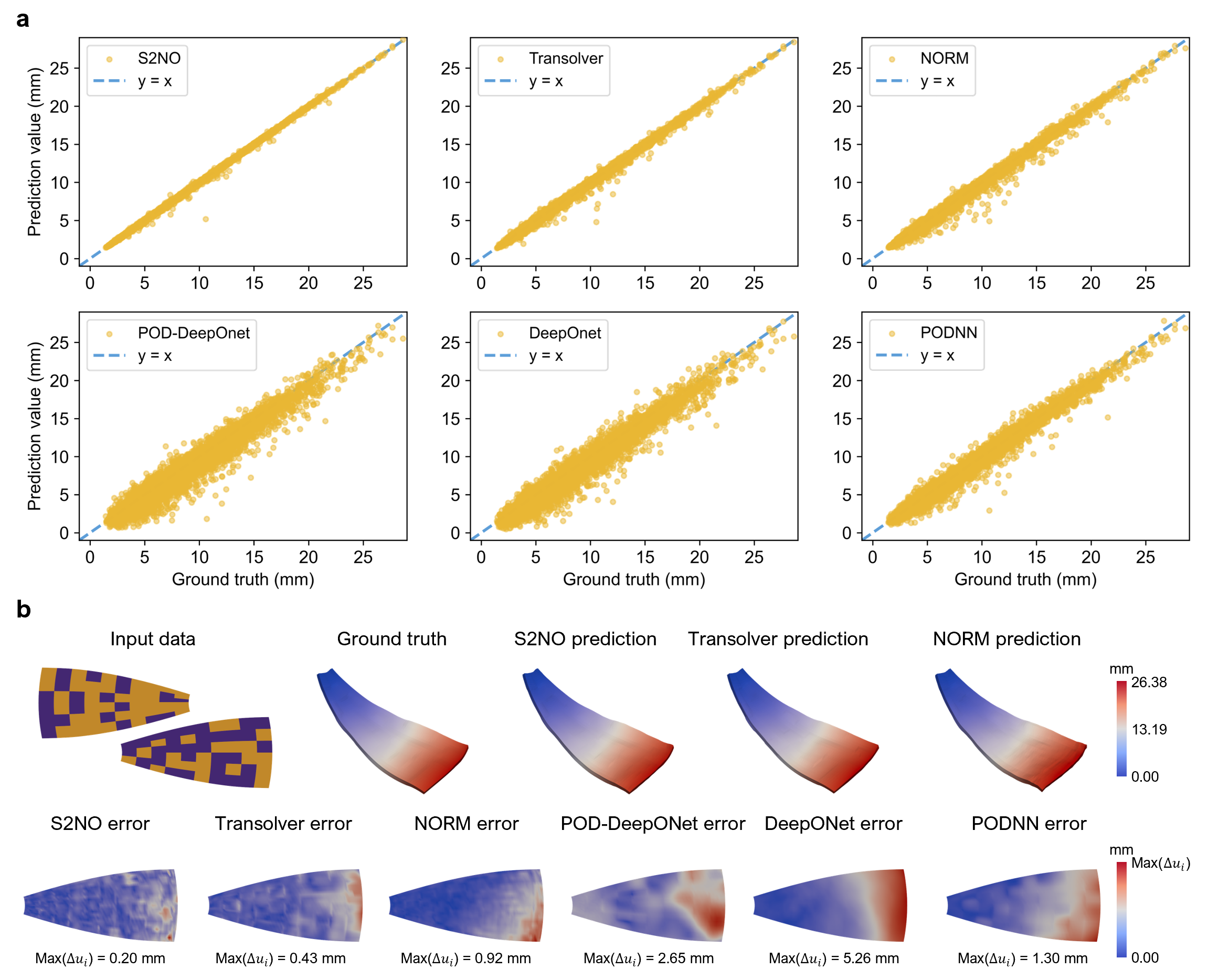}
	\caption{Prediction results of the S2NO model and baseline models for the blade case. \textbf{a}, Scatter plot of prediction values versus ground truth of the maximum deformation for each test sample. Results closer to the reference line $y=x$ indicate higher predictive accuracy of the model. \textbf{b}, Input material distribution, ground truth, prediction results, and error maps for a data sample randomly selected from the test dataset are presented to compare the predictive performance of different models visually. $\operatorname{Max}\left(\Delta u_i\right)$ represents the maximum value of the point-to-point distance between the predicted shape and the ground truth. The $\operatorname{Max}\left(\Delta u_i\right)$ of S2NO is merely 0.20 mm.}
	\label{fig:fig10}
\end{figure}

\clearpage
\begin{figure}
	\centering
	\includegraphics[width=1.0\linewidth]{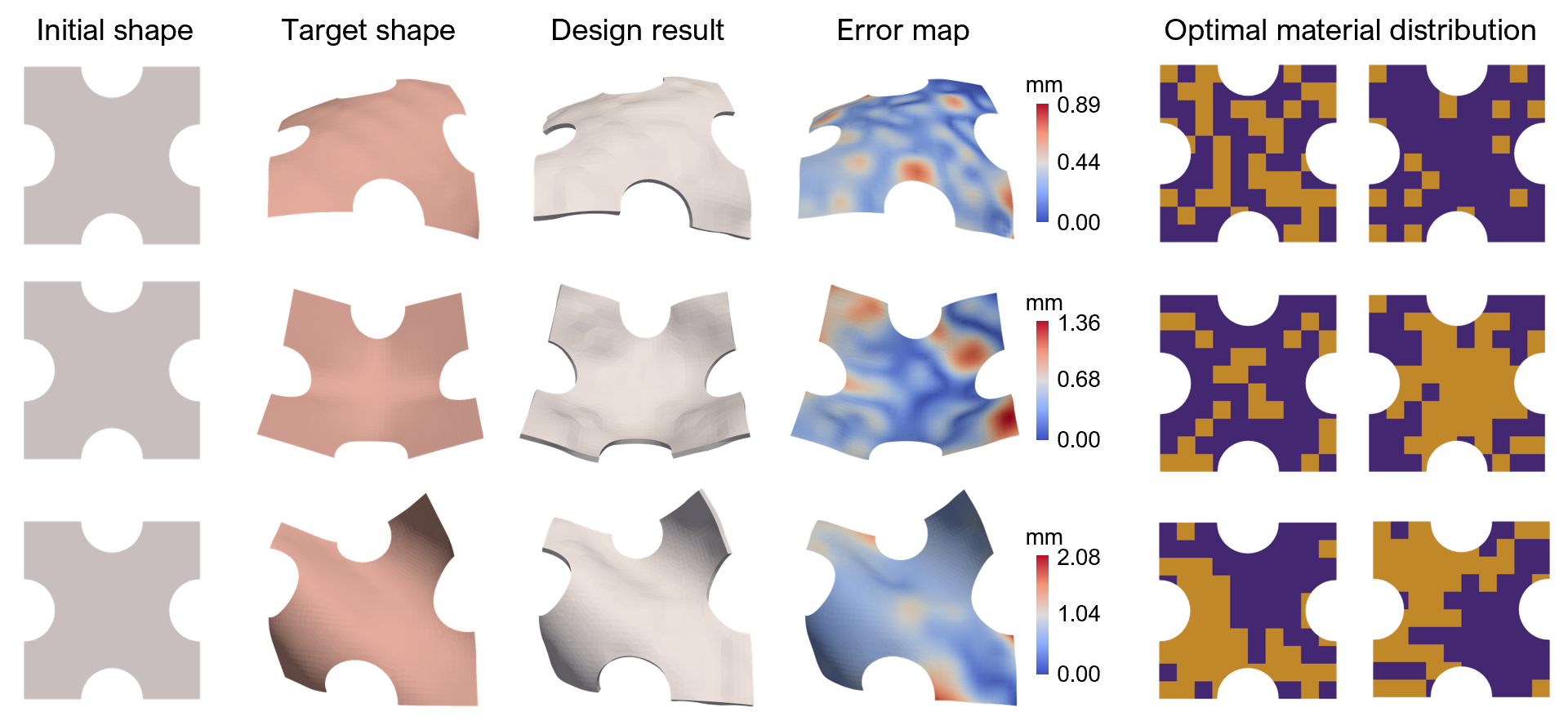}
	\caption{Initial shapes, target shapes, design results, error maps between the target and the design, and the corresponding optimal material distributions for the dart case.}
	\label{fig:fig11}
\end{figure}

\begin{figure}
	\centering
	\includegraphics[width=1.0\linewidth]{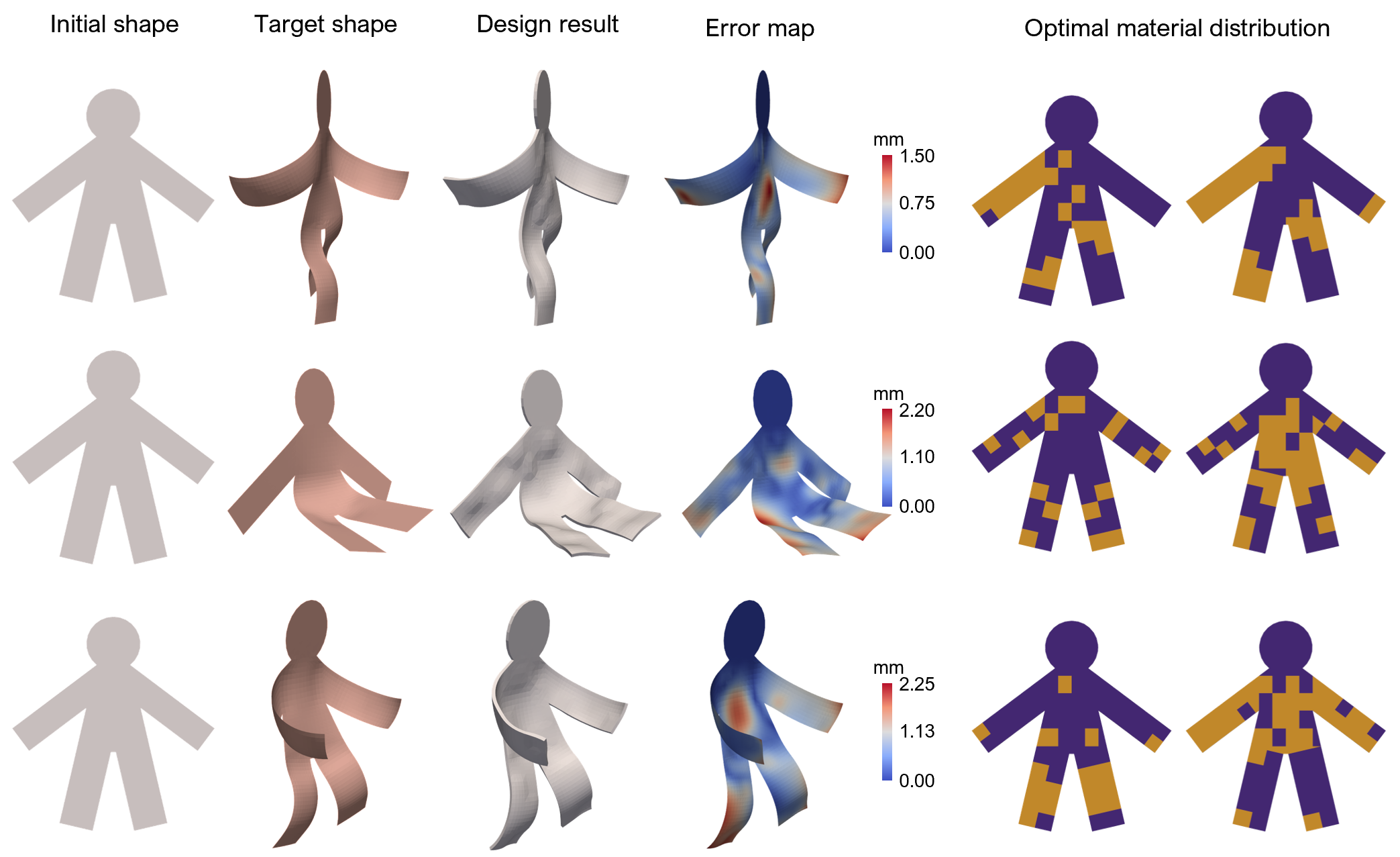}
	\caption{Initial shapes, target shapes, design results, error maps between the target and the design, and the corresponding optimal material distributions for the human case.}
	\label{fig:fig12}
\end{figure}

\clearpage
\begin{figure}
	\centering
	\includegraphics[width=1.0\linewidth]{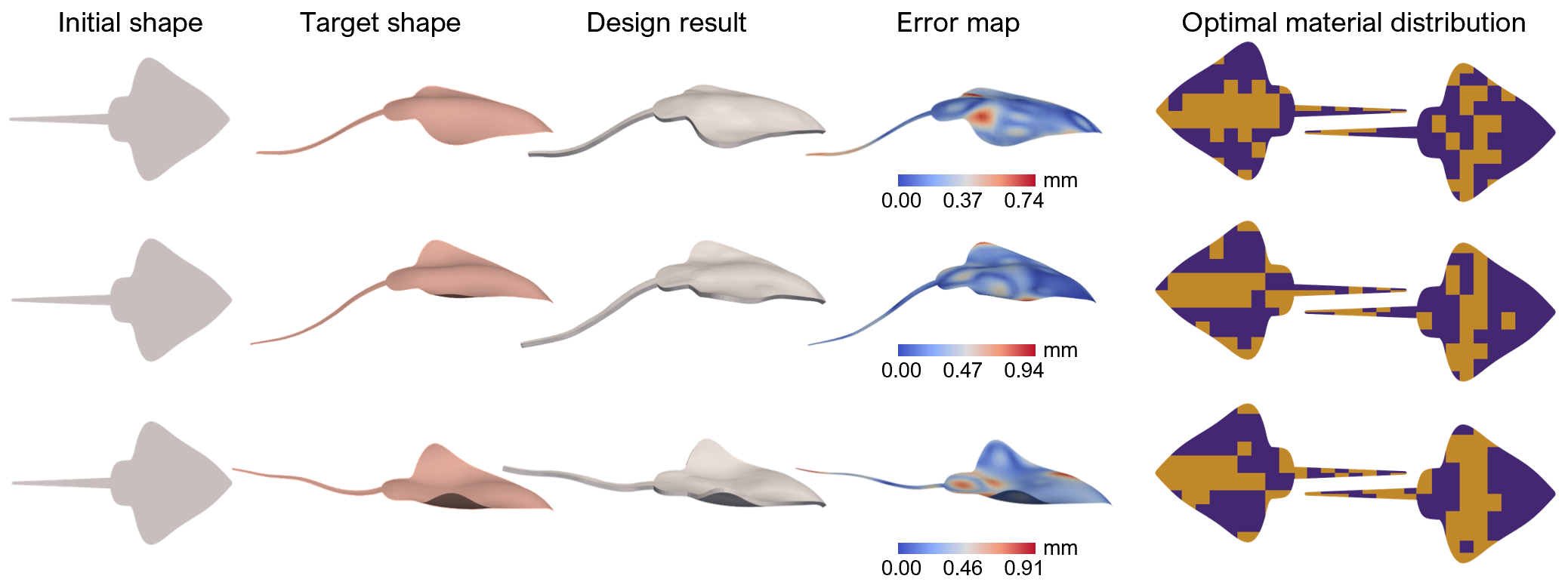}
	\caption{Initial shapes, target shapes, design results, error maps between the target and the design, and the corresponding optimal material distributions for the stingray case.}
	\label{fig:fig13}
\end{figure}

\begin{figure}
	\centering
	\includegraphics[width=1.0\linewidth]{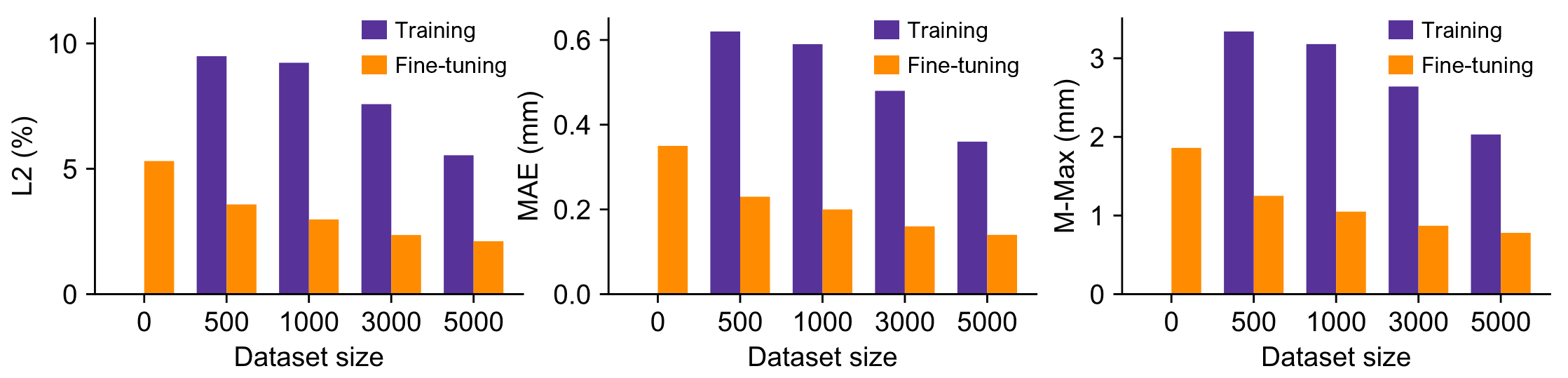}
	\caption{Performance comparison on the dart case of model fine-tuning versus direct training from scratch on limited data and direct predictions from the low-resolution model (Dataset size is 0).}
	\label{fig:fig14}
\end{figure}

\clearpage
\begin{figure}
	\centering
	\includegraphics[width=1.0\linewidth]{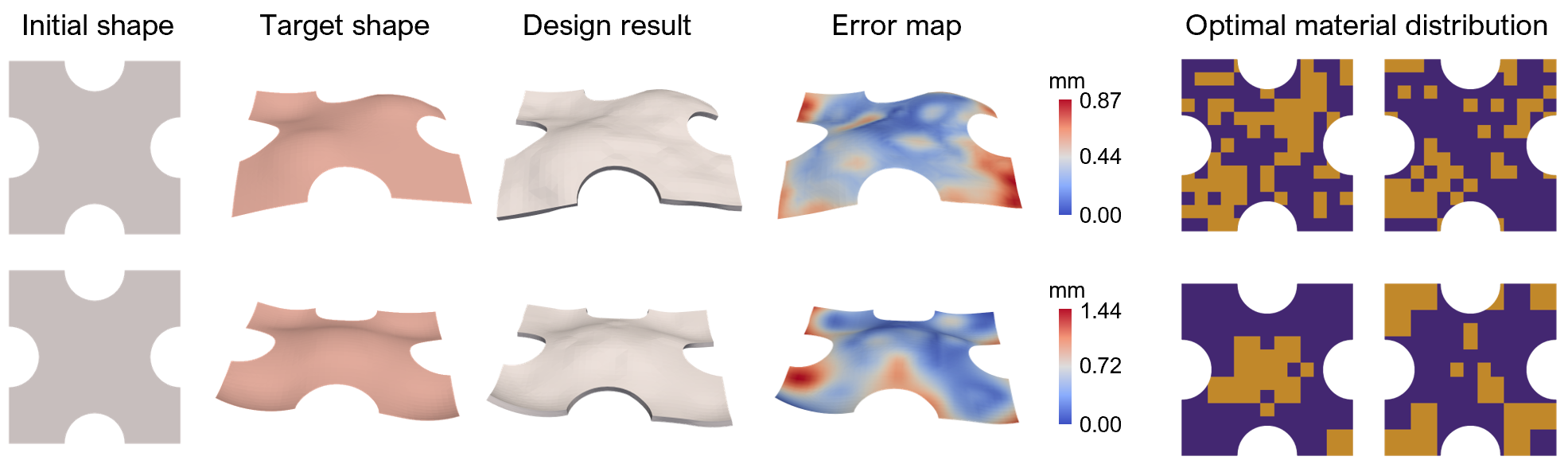}
	\caption{Initial shapes, target shapes, design results, error maps between the target and the design, and the corresponding optimal material distributions for the high-resolution dart case.}
	\label{fig:fig15}
\end{figure}

\begin{figure}
	\centering
	\includegraphics[width=1.0\linewidth]{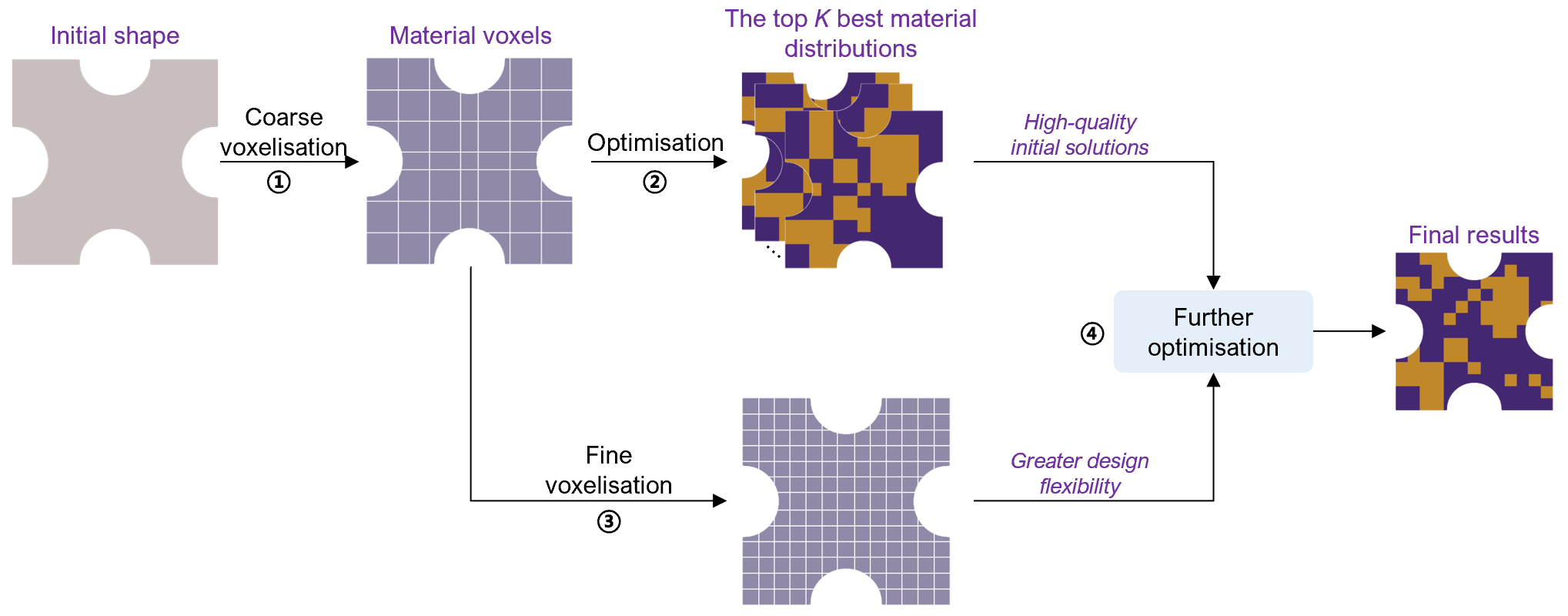}
	\caption{A comprehensive process for the multi-resolution optimisation strategy. The initial shape is first coarsely voxelised and optimised to generate high-quality initial solutions. These solutions then serve as the starting point for further optimisation after fine voxelisation, yielding the final high-resolution design.}
	\label{fig:fig16}
\end{figure}

\begin{figure}
	\centering
	\includegraphics[width=1.0\linewidth]{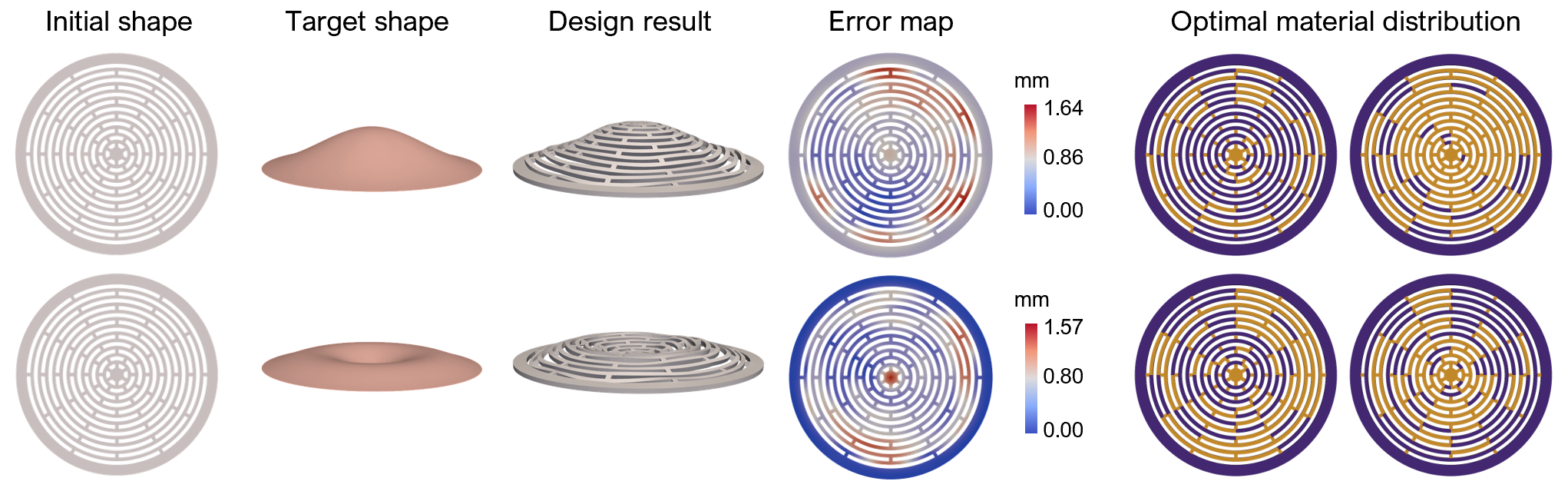}
	\caption{Initial shapes, target shapes, design results, error maps between the target and the design, and the corresponding optimal material distributions for the dome case.}
	\label{fig:fig17}
\end{figure}

\clearpage
\begin{figure}
	\centering
	\includegraphics[width=1.0\linewidth]{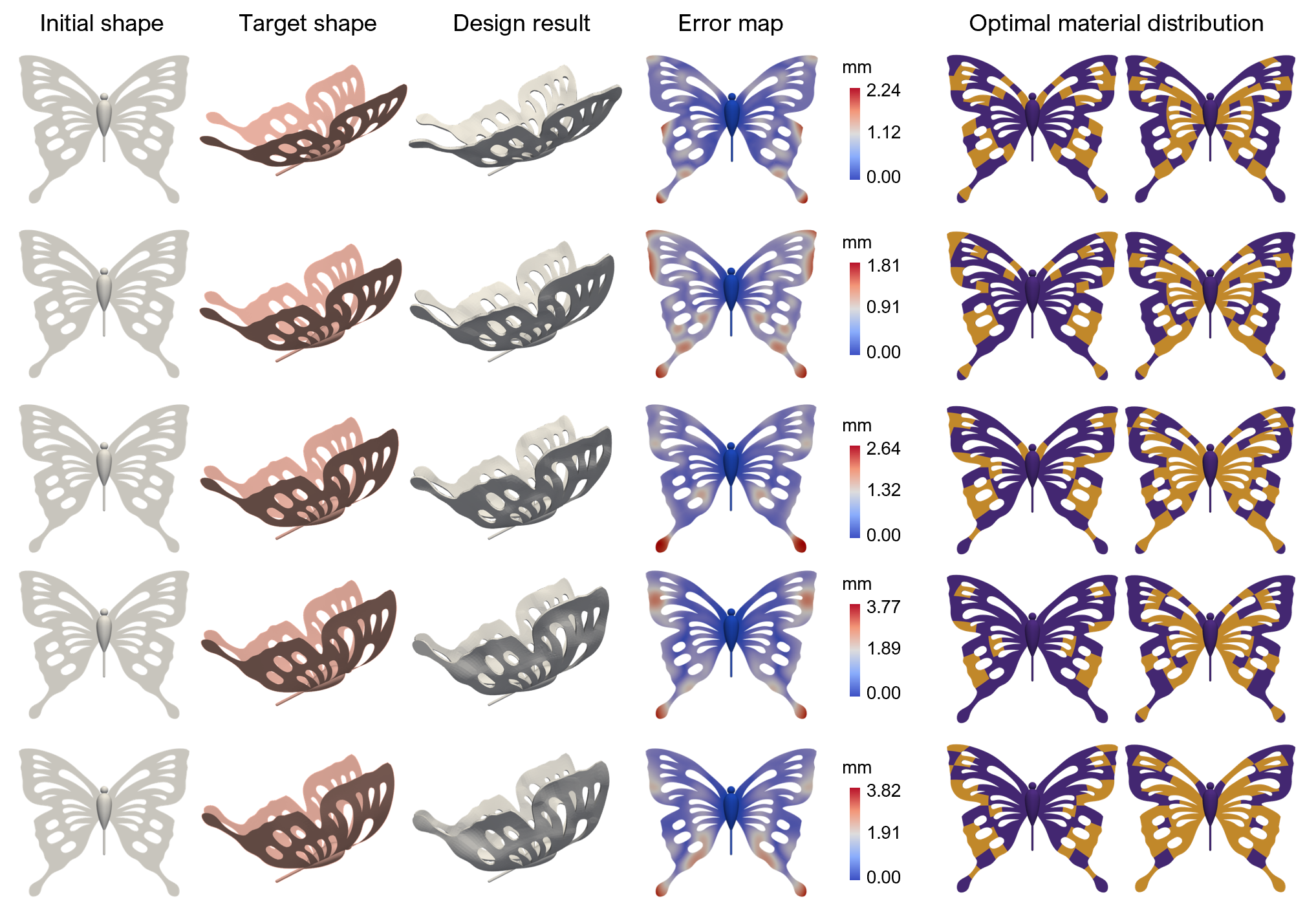}
	\caption{Initial shapes, target shapes, design results, error maps between the target and the design, and the corresponding optimal material distributions for the butterfly case.}
	\label{fig:fig18}
\end{figure}

\begin{figure}
	\centering
	\includegraphics[width=1.0\linewidth]{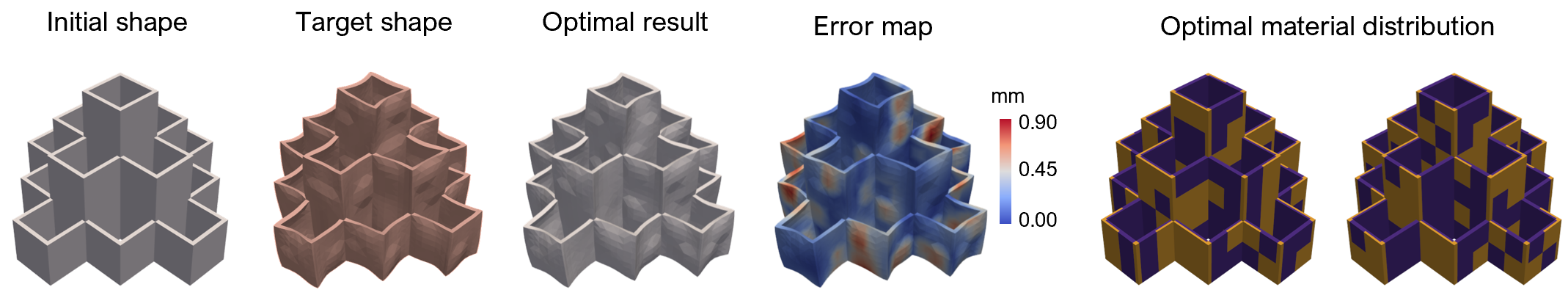}
	\caption{Initial shapes, target shapes, design results, error maps between the target and the design, and the corresponding optimal material distributions for the 3D thin-walled structure.}
	\label{fig:fig19}
\end{figure}

\clearpage
\begin{figure}
	\centering
	\includegraphics[width=1.0\linewidth]{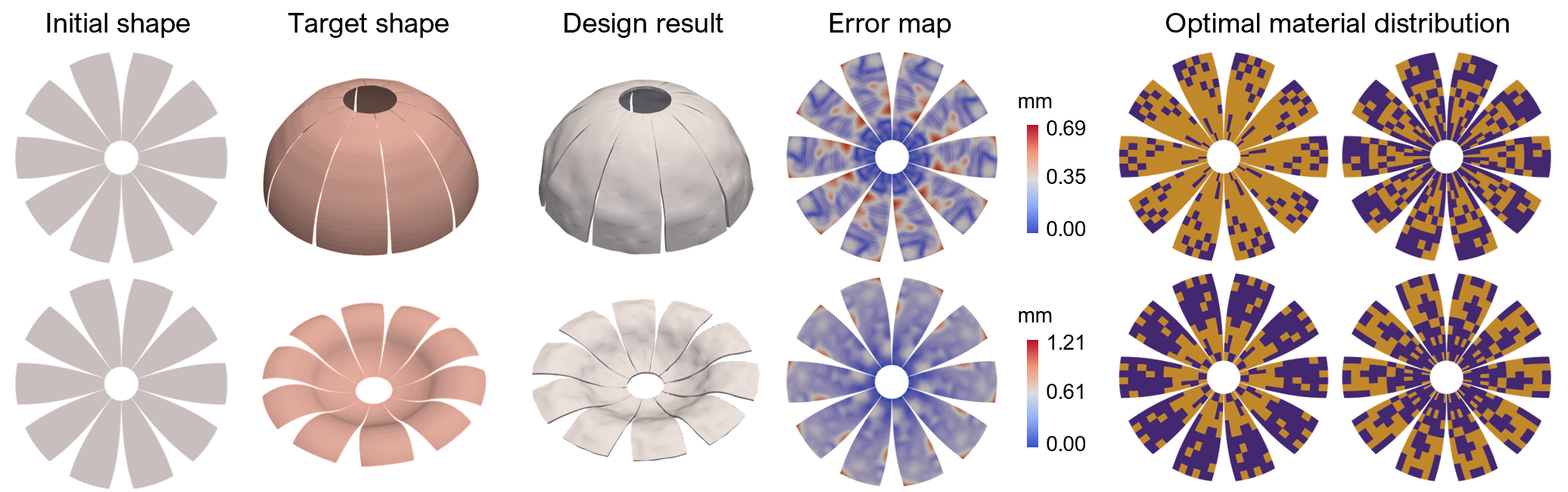}
	\caption{Initial shapes, target shapes, design results, error maps between the target and the design, and the corresponding optimal material distributions for the blade case.}
	\label{fig:fig20}
\end{figure}

\begin{figure}
	\centering
	\includegraphics[width=1.0\linewidth]{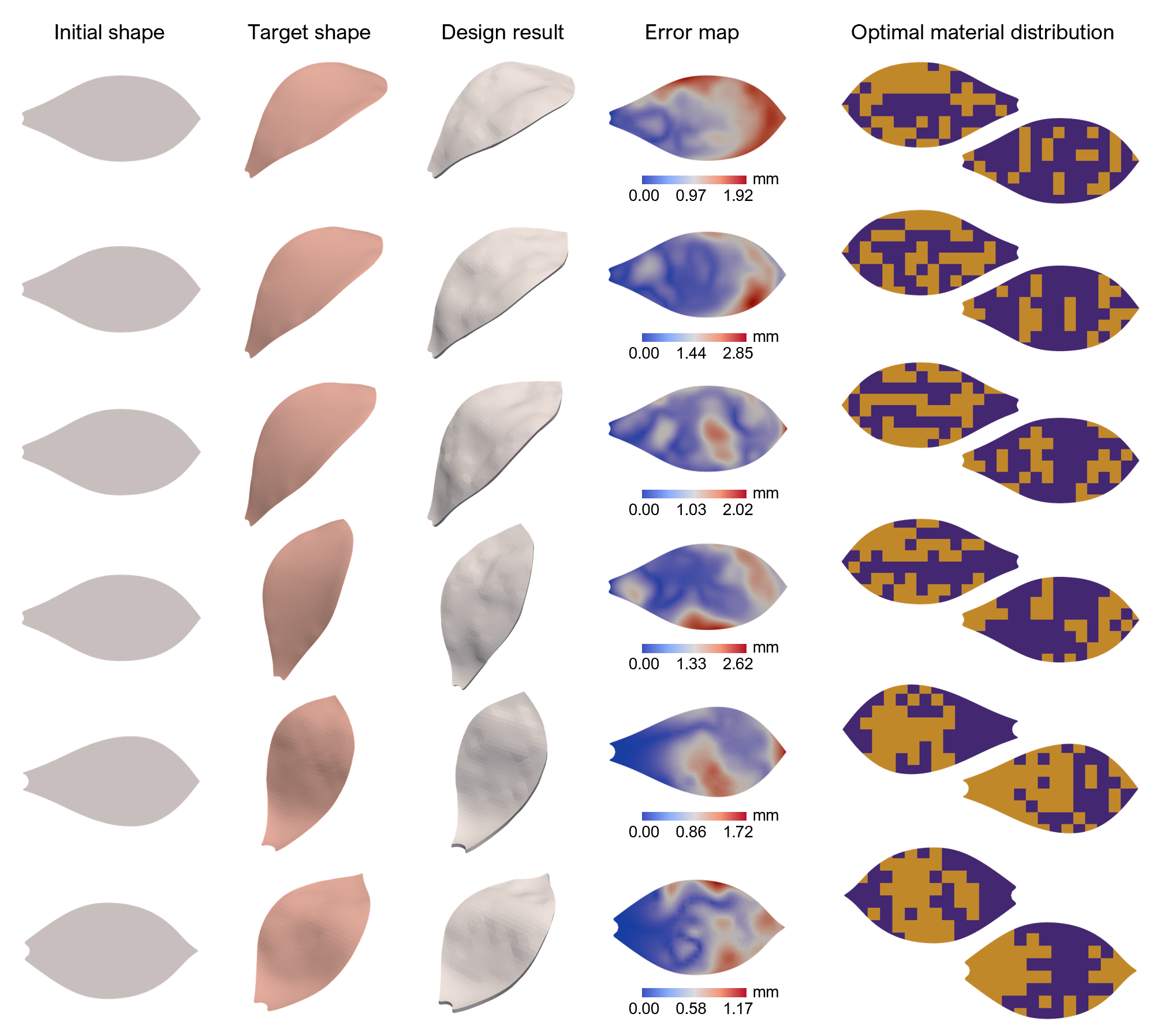}
	\caption{Initial shapes, target shapes, design results, error maps between the target and the design, and the corresponding optimal material distributions for the petal case.}
	\label{fig:fig21}
\end{figure}

\clearpage
\section{Supplementary Tables}

\begin{table}[htbp]
	
	\centering
	\caption{Implementation detail for each data-driven model.}
	\label{tab:S1}
	\footnotesize
	\setlength{\tabcolsep}{3pt}
	\begin{tabular}{@{}>{\raggedright\arraybackslash}m{1.8cm}ccccccccc@{}}
		\toprule
		& \multicolumn{4}{c}{Model configurations} & \multicolumn{5}{c}{Training configurations} \\
		\cmidrule(lr){2-5} \cmidrule(lr){6-10}
		Model & Layers & Modes & Channels & Heads & Optimiser & Scheduler & \makecell{Initial learning rate} & Epochs & \makecell{Batch size} \\
		\midrule
		S2NO & 8 & 128 & 128 & 8 & AdamW & OneCycleLR & 1e-3 & 500 & 16 \\
		\midrule
		Transolver & 8 & 32 & 128 & 8 & AdamW & OneCycleLR & 1e-3 & 500 & 16 \\
		\midrule
		NORM & 5 & 128 & 64 & / & AdamW & OneCycleLR & 1e-3 & 500 & 16 \\
		\midrule
		POD-DeepONet & \multicolumn{4}{l}{\makecell[l]{POD modes: 128\\Branch net: $[n, 512, 256, 256, 128]$}} & Adam & StepLR & 1e-3 & 2000 & 200 \\
		\midrule
		DeepONet & \multicolumn{4}{l}{\makecell[l]{Branch net: $[n, 512, 256, 256, 128]$\\Trunk net: $[3, 128, 256, 256, 128]$}} & Adam & StepLR & 1e-3 & 2000 & 200 \\
		\midrule
		PODNN & \multicolumn{4}{l}{\makecell[l]{POD modes: 64\\NN architecture: $[m, 512, 512, 512, 512, 64]$}} & Adam & StepLR & 1e-3 & 2000 & 200 \\
		\bottomrule
	\end{tabular}
	\\[6pt]
	\parbox{\linewidth}{\footnotesize
		``Layers'' represent the number of L-layers in NORM, S2NO-layers in S2NO, or blocks in Transolver. ``Modes'' represent the number of Laplacian eigenfunctions in S2NO and NORM or the number of slices in the physical attention mechanism in Transolver. ``Channels'' represent the dimensions after lifting. ``Heads'' represent the number of attention heads in a multi-head architecture. $n$ represents the number of discretisation points on the geometry, and $m$ represents the number of designable material voxels.}
\end{table}

\vspace{8em} 
\begin{table}[htbp]
	\centering
	\caption{Performance comparison on the dart case of model fine-tuning versus direct training from scratch on limited data.}
	\label{tab:S2}
	\footnotesize
	\begin{tabularx}{\textwidth}{@{}XXXX@{}}
		\toprule
		Dataset size & Metric & Training & Fine-tuning\\
		\midrule
		\multirow{3}{*}{500} 
		& L2 (\%) & 9.49 (3.1e-1) & \textbf{3.58 (4.3e-3)} \\
		& MAE (mm) & 0.62 (2.2e-2) & \textbf{0.23 (2.7e-4)} \\
		& M-Max (mm) & 3.34 (7.6e-2) & \textbf{1.25 (3.5e-3)} \\
		\midrule
		\multirow{3}{*}{1000} 
		& L2 (\%) & 9.23 (1.5e-1) & \textbf{2.98 (8.8e-4)} \\
		& MAE (mm) & 0.59 (1.1e-2) & \textbf{0.20 (5.8e-5)} \\
		& M-Max (mm) & 3.18 (4.8e-2) & \textbf{1.05 (1.9e-3)} \\
		\midrule
		\multirow{3}{*}{3000} 
		& L2 (\%) & 7.58 (2.7e-1) & \textbf{2.36 (1.8e-3)} \\
		& MAE (mm) & 0.48 (1.5e-2) & \textbf{0.16 (9.3e-5)} \\
		& M-Max (mm) & 2.64 (7.5e-2) & \textbf{0.87 (2.2e-3)} \\
		\midrule
		\multirow{3}{*}{5000} 
		& L2 (\%) & 5.55 (2.9e-1) & \textbf{2.11 (2.6e-3)} \\
		& MAE (mm) & 0.36 (1.9e-2) & \textbf{0.14 (1.7e-4)} \\
		& M-Max (mm) & 2.03 (9.4e-2) & \textbf{0.78 (1.0e-3)} \\
		\bottomrule
	\end{tabularx}
\end{table}	

\clearpage
\begin{table}[htbp]
	\centering
	\caption{Performance comparison on five petals of S2NO-based multi-geometry modelling versus separate training.}
	\label{tab:S3}
	\footnotesize
	\begin{tabularx}{\textwidth}{@{}XXXX@{}}
		\toprule
		Case & Metric & Separate training & Multi-geometry modelling \\
		\midrule
		\multirow{3}{*}{Petal A} 
		& L2 (\%) & 2.55 (1.0e-1) & \textbf{1.61 (5.5e-2)} \\
		& MAE (mm) & 0.35 (1.6e-2) & \textbf{0.22 (8.2e-3)} \\
		& M-Max (mm) & 2.52 (9.4e-2) & \textbf{1.63 (5.2e-2)} \\
		\midrule
		\multirow{3}{*}{Petal B} 
		& L2 (\%) & 1.05 (2.6e-2) & \textbf{0.78 (2.5e-2)} \\
		& MAE (mm) & 0.09 (2.5e-3) & \textbf{0.07 (2.3e-3)} \\
		& M-Max (mm) & 0.77 (1.5e-2) & \textbf{0.57 (1.6e-2)} \\
		\midrule
		\multirow{3}{*}{Petal C} 
		& L2 (\%) & 0.87 (2.0e-2) & \textbf{0.75 (3.3e-2)} \\
		& MAE (mm) & 0.07 (1.0e-3) & \textbf{0.06 (2.7e-3)} \\
		& M-Max (mm) & 0.59 (1.2e-2) & \textbf{0.53 (2.1e-2)} \\
		\midrule
		\multirow{3}{*}{Petal D} 
		& L2 (\%) & 1.26 (1.9e-2) & \textbf{0.91 (4.6e-2)} \\
		& MAE (mm) & 0.10 (2.1e-3) & \textbf{0.08 (3.7e-3)} \\
		& M-Max (mm) & 0.77 (1.7e-2) & \textbf{0.56 (2.4e-2)} \\
		\midrule[\heavyrulewidth]  
		\multirow{3}{*}{Petal E} 
		& L2 (\%) & 1.74 (2.2e-2) & \textbf{1.13 (4.2e-2)} \\
		& MAE (mm) & 0.16 (1.5e-3) & \textbf{0.11 (4.2e-3)} \\
		& M-Max (mm) & 1.26 (2.2e-2) & \textbf{0.84 (2.7e-2)} \\
		\bottomrule
	\end{tabularx}
\end{table}	

\bibliographystyle{unsrtnat}
\bibliography{references_supp}  




